\documentclass[11pt]{article}

\usepackage[preprint]{acl}

\usepackage{times}
\usepackage{latexsym}

\usepackage[T1]{fontenc}

\usepackage[utf8]{inputenc}

\usepackage{microtype}

\usepackage{inconsolata}

\usepackage{graphicx}
\usepackage{algorithm}
\usepackage{algorithmic}
\usepackage{array} 
\usepackage{amsmath}
\usepackage{algorithmic}
\usepackage{multirow}
\usepackage{amsmath}
\usepackage{xcolor}
\usepackage{caption}
\usepackage{enumitem}
\usepackage{adjustbox}
\usepackage{multirow}
\usepackage{mathtools}

\usepackage{float}
\usepackage{booktabs}

\title{Intention-Adaptive LLM Fine-Tuning for Text Revision Generation}

\author{Zhexiong Liu, Diane Litman
 \\
    Department of Computer Science, 
  Learning Research \& Development Center \\
University of Pittsburgh, Pittsburgh, Pennsylvania, USA 15260
\\
   \texttt{zhexiong@cs.pitt.edu}, \texttt{dlitman@pitt.edu}
  }

\begin{document}
\maketitle
\begin{abstract}
Large Language Models (LLMs) have achieved impressive capabilities in various context-based text generation tasks, such as summarization and reasoning; however, their applications in intention-based generation tasks remain underexplored. One such example is revision generation, which requires the generated text to explicitly reflect the writer's actual intentions. Identifying intentions and generating desirable revisions are challenging due to their complex and diverse nature. Although prior work has employed LLMs to generate revisions with few-shot learning, they struggle with handling entangled multi-intent scenarios. While fine-tuning LLMs using intention-based instructions appears promising, it demands large amounts of annotated data, which is expensive and scarce in the revision community. To address these challenges, we propose Intention-Tuning, an intention-adaptive layer-wise LLM fine-tuning framework that dynamically selects a subset of LLM layers to learn the intentions and subsequently transfers their representations to revision generation. Experimental results suggest that Intention-Tuning is effective and efficient on small revision corpora, outperforming several PEFT baselines.
\end{abstract}

\section{Introduction}
\begin{figure}[tb]
    \centering
    \includegraphics[width=0.8\columnwidth]{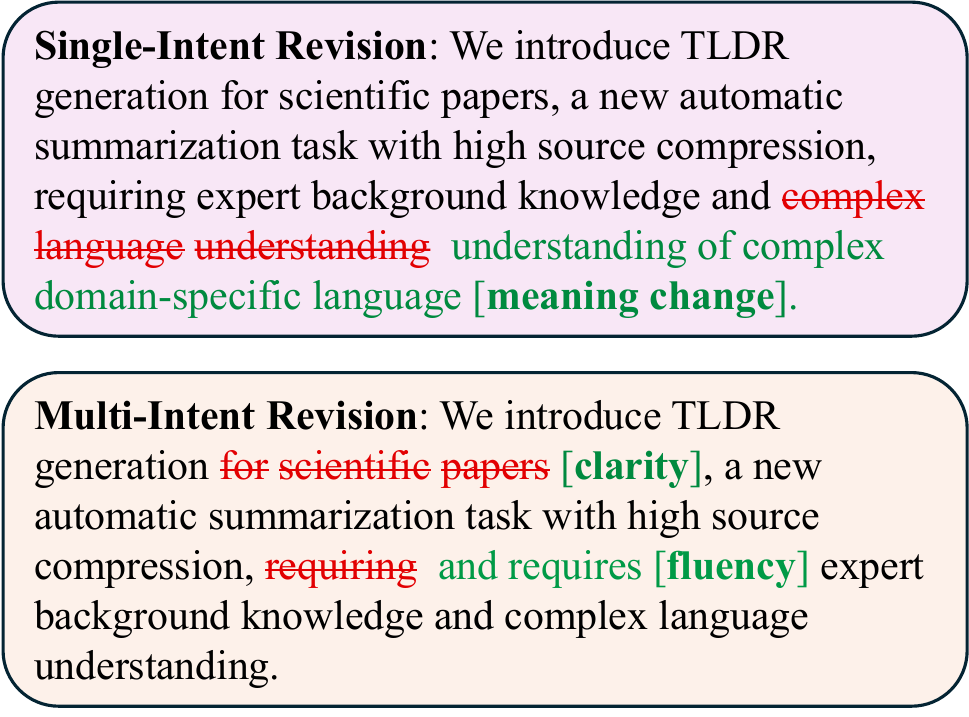}
    \vspace{-.072in}
    \caption{Revision examples: the original text is revised differently based on a single (meaning change) intention and multiple (clarity and fluency) intentions. The examples are from the ITERATER corpus~\cite{du-etal-2022-understanding-iterative}.}
    \vspace{-.168in}
    \label{fig:intro-example}
\end{figure}

Text revision has been regarded as an essential part of writing as it typically improves the final written work through multiple rounds of editing~\cite{sommers1980revision}. However, a writer's actual intentions, which motivate the revision, are difficult to capture due to their entangled and multi-intent nature~\cite{fitzgerald1987research}. Figure~\ref{fig:intro-example} shows the same text edited based on single-intent and multi-intent revisions. One involves a \textit{meaning-changed} revision, while the other focuses on \textit{clarity} and \textit{fluency} improvement. These examples demonstrate that text revisions are often driven by diverse intentions, whether single or multiple, which makes it challenging to generate revisions using computational methods. In the literature, prior work primarily employs sequence-to-sequence language models, such as BART~\cite{lewis2019bart}, to generate revisions while keeping intentions as inputs~\cite{du-etal-2022-read,du-etal-2022-understanding-iterative}, or utilizes prefix-tuning to learn intention representations for revision generation~\cite{chong-etal-2023-leveraging}. However, these models often struggle with multiple entangled intentions, due to their limited capacity~\cite{skitalinskaya-wachsmuth-2023-revise}. Although~\citet{ziegenbein-etal-2024-llm} utilize reinforcement learning policies to prompt large language models (LLMs) for argument rewriting, no specific intentions are used for fine-tuning. While~\citet{shu2024rewritelm} develop LLMs for text rewriting, their fine-tuning is based on implicit writing instructions rather than explicit intentions. We argue that LLMs fine-tuned on dedicated intention-based revision corpora are critically needed for addressing revision tasks.

LLMs have achieved impressive success in various NLP tasks, including summarization~\cite{takeshita-etal-2024-aclsum}, reasoning~\cite{li2024cr}, and question answering~\cite{peng2024chain}. However, their application in revision tasks remains underexplored. This might be because text revision requires an iterative editing process, involving additions, deletions, and modifications, each driven by specific intentions; nevertheless, LLMs are mostly pre-trained to generate just final texts. Although~\citet{shu2024rewritelm,ruan-etal-2024-large} explore revision LLMs using parameter-efficient fine-tuning (PEFT) methods, these approaches require large-scale training data. In contrast, most revision corpora are small and contain limited annotations, making such methods unsustainable in the revision community.

Increasing attention has recently been given to layer-wise PEFT, since prior work~\cite{zhang-etal-2023-crash,elhoushi-etal-2024-layerskip} suggests fine-tuning LLM layers that contribute more, while freezing those that contribute less, could benefit downstream tasks. For instance,~\citet{yao-etal-2024-layer} introduce an importance-aware sparse tuning (IST) to sample a subset of LLM layers for PEFT. While effective, it suffers from over-sampling or under-sampling issues, resulting in suboptimal results~\cite{liu-litman-2025-ir-tuning}. In addition, current layer-wise PEFT methods~\cite{pan2024lisa,zhu2023lift,yao-etal-2024-layer,wei-etal-2025-flexora} have limited exploration in revision generation tasks, since they are primarily fine-tuned on context-based, which relies heavily on contextualized understanding, rather than intention-based generation, which requires the revisions to specifically reflect the writer’s actual intentions. Therefore, we propose Intention-Tuning\footnote{\href{https://github.com/ZhexiongLiu/Intention-Tuning}{https://github.com/ZhexiongLiu/Intention-Tuning}}, a novel intention-adaptive PEFT framework dedicated to learning revision intentions in one task and fine-tuning revision generation in another, all through selected LLM layers. Although Intention-Tuning introduces an additional task to facilitate the transfer of learned representations, it remains efficient as the two tasks are fine-tuned in a sequential manner rather than a joint optimization. 

To assess the Intention-Tuning for revision generation, we study three research questions: \textbf{RQ1}: Can Intention-Tuning align intention prediction and revision generation tasks through selected LLM layers? \textbf{RQ2}: Can Intention-Tuning generate effective revisions using small annotated corpora? \textbf{RQ3}: Can Intention-Tuning be generally efficient across different PEFT adapters? In particular, we make the following contributions:

\begin{itemize}
[leftmargin=*,itemsep=-5pt, topsep=0pt]
\item We are the first work to use an intention-adaptive layer-wise PEFT method for revision generation.
\item We develop a framework to align revision intention and revision generation through LLM layers.
\item We demonstrate the feasibility and generalizability of the framework on small revision corpora.
\end{itemize}

\section{Related Work}
\paragraph{Revision Generation} 
Text revision primarily focuses on analyzing human edits to identify revision intentions and generating revisions based on the specific intentions~\cite{skitalinskaya-wachsmuth-2023-revise,skitalinskaya-etal-2023-claim,mita-etal-2024-towards,jourdan-etal-2024-casimir,ruan-etal-2024-large}. Although an essential NLP task, revision generation remains challenging, mainly due to the scarcity of human annotated corpora~\cite{anthonio-etal-2020-wikihowtoimprove,spangher-etal-2022-newsedits,du-etal-2022-understanding-iterative,darcy-etal-2024-aries,liu-etal-2025-erevise}. Prior work~\cite{afrin2018improvement, kashefi2022argrewrite,afrin2020RER,afrin2023predicting} develops feature-based approaches to predict revision intentions but not revision generation. While~\citet{arxivedits,chong-etal-2023-leveraging} utilize sequence-to-sequence language models, such as BART~\cite{lewis2019bart}, to study revision generation, these models struggle to capture complex revision patterns due to their limited capacity. More recent efforts have been made to prompt LLMs using instructional inputs~\cite{ruan-etal-2024-re3} and fine-tune LLMs~\cite{ziegenbein-etal-2024-llm,shu2024rewritelm} on a large amount of data. In contrast, we explore more efficient layer-wise PEFT with small corpora.

\paragraph{Layer-wise PEFT}
PEFT provides efficient solutions for LLM fine-tuning, as it employs adapter-based~\cite{hu-etal-2023-llm}, low-rank adaptive~\cite{hu2022lora}, and prompt-based tuning~\cite{zhao-etal-2024-layer} to minimize expense. However, these methods apply an identical PEFT strategy across all LLM layers, which cannot leverage distinct layer-wise contributions to downstream tasks. To address it,~\citet{kaplun2023less} introduce a greedy search to select informative layers, \citet{pan2024lisa} instead adapt random layer selection, and \citet{zhu2023lift} use directional heuristics for the layer-wise fine-tuning. Despite their effectiveness, these methods either incur high computational costs or rely on simple heuristics incompatible with complex tasks. In addition, \citet{yao-etal-2024-layer,wei-etal-2025-flexora} propose sampling important layers based on scoring metrics; however, their methods risk sampling too many redundant layers due to their fixed number of layer selections. Although~\citet{liu-litman-2025-ir-tuning} address this issue using a dynamic method, they focus on intention prediction rather than revision generation.
Recently, PEFT has been integrated with multi-task learning.~\citet{liu2024mftcoder} fine-tune LLMs by leveraging LoRA~\cite{hu2022lora} and QLoRA~\cite{qlora2023} to optimize a balanced multi-task approach, while~\citet{agiza2024mtlora} employ task-agnostic and task-specific PEFT adapters for joint fine-tuning. In addition,~\citet{baek2025tadformer} develop task-aware feature adaptation by considering task-specific input contexts, whereas~\citet {cheng2025compmtl} mitigate task-wise gradient conflicts at individual model layers. Although helpful, these methods primarily focused on either shared adapter modules or non-LLM layers. In contrast, we build on prior work~\cite{liu-litman-2025-ir-tuning} to dynamically select LLM layers for intention-adaptive multi-task PEFT.

\begin{figure}[tb]
    \centering
    \includegraphics[width=1\columnwidth]{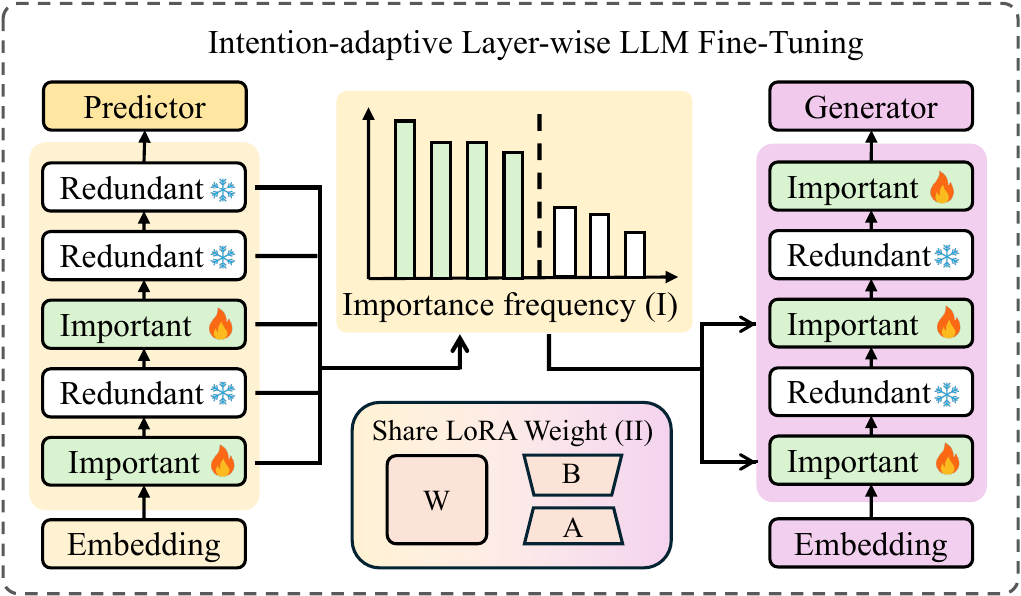}
    \vspace{-.25in}
    \caption{The Intention-Tuning framework. In the intention prediction task (predictor), the important LLM layers are fine-tuned while the redundant LLM layers are frozen. Upon completion, (I) its layer-wise importance frequency is used to finalize the important layers for the revision generation task (generator); (II) the learned layer-wise representations (LoRA weights) in the predictor are shared with the generator.} 
    \label{fig:framework}
    \vspace{-.15in}
\end{figure}

\section{Methods}
\subsection{Preliminaries}
We use the notations based on the previous text revision tasks~\cite{liu-etal-2023-predicting}. 
We denote $\mathcal{X}$, $\mathcal{Y}$, and $\mathcal{I}$ as original text, revised text, and revision intentions, respectively. We formulate revision intention prediction task $\mathcal{T}_{pre}=f(\mathcal{I;\mathcal{X}, \mathcal{Y}})$ as classifying a pair of revision $\{\mathcal{X}, \mathcal{Y}\}$ into an intention label $\mathcal{I}$, and revision generation task $\mathcal{T}_{gen}=f(\mathcal{Y;\mathcal{X}, \mathcal{I}})$ as generating $\mathcal{Y}$ given a pair of the original text and its revision intention $\{\mathcal{X}, \mathcal{I}\}$. We propose an intention-adaptive layer-wise PEFT framework that learns intentions in one task and generates revisions in another, all through selected LLM layers. Figure~\ref{fig:framework} shows the Intention-Tuning framework.

\subsection{LLM Layer Selection}
Inspired by work~\cite{zhang-etal-2023-crash,elhoushi-etal-2024-layerskip,wei-etal-2025-flexora} that suggests fine-tuning partial LLM layers is a feasible PEFT strategy, we use a two-step method to implement this idea, which first probes important layers in $\mathcal{T}_{pre}$, and then uses the importance frequency based on the probed layers to finalize the important layers in $\mathcal{T}_{gen}$. Note that the used layers in both tasks have high overlap but might not be exactly the same.

\label{sec:layer_selection}
\paragraph{Probing Important Layer} Given an LLM $\mathcal{M}=\left\{m_i\right\}_{i=1}^{\ell}$ consisting of $\ell$ transformer layer $m$. The objective is to split $\ell$ layers into an important subset $\mathcal{S}$ and a redundant subset $\bar{\mathcal{S}}$, where the layers in $\mathcal{S}$ are used for fine-tuning, and the layers in $\bar{\mathcal{S}}$ are frozen. We use the method in~\citet{liu-litman-2025-ir-tuning} to probe important layers based on the layer-wise importance scores (gradient norms) in $\mathcal{M}$ for the task $\mathcal{T}_{pre}$, as they suggest that layer parameters with high gradient norms make key contributions to the rapid update of the loss function, thus facilitating efficient gradient descent. Additionally, layers with large gradient norms carry information relevant to downstream tasks, making LLM layers informative during fine-tuning. The detailed probing method is described in Appendix~\ref{sec:ir_method}.  

\paragraph{Finalizing Important Layer} 
Suppose the task $\mathcal{T}_{pre}$ obtains an important layer subset $\mathcal{S}_{pre}$ and redundant layer subset $\bar{\mathcal{S}}_{pre}$ of $\mathcal{M}$ in a single fine-tuning step, its layer-wise importance (updating) frequency $\mathcal{G}=\{p_i\}_{i=1}^{\ell}$ can be obtained, where $p_i$ is the frequency of transformer layer $m_i$ being selected throughout a $k$-step fine-tuning. Here, $p_i=\sum_{j=1}^{k}q_{i,j}$, where 
\begin{align}
q_{i,j} &= 
\begin{cases}
\label{eq:frequency}
1 & \text{if layer } m_i \text{ is selected at step $j$,} \\
0 & \text{otherwise.}
\end{cases}
\end{align}
\noindent We argue that frequently selected layers in task $\mathcal{T}_{pre}$ are well-trained to capture revision intentions, which can be reused for the task of revision generation $\mathcal{T}_{gen}$. Specifically, layer-wise updating frequency $\mathcal{G}$ from task $\mathcal{T}_{pre}$ is again used to split $\ell$ layers of $\mathcal{M}$ into important subset $\mathcal{S}_{gen}$ and redundant subset $\bar{\mathcal{S}}_{gen}$ for the task of $\mathcal{T}_{gen}$, by solving Equation~\ref{eq:likelihood},~\ref{eq:variance} in the Appendix. This design ensures learned representations in the revision intention task can be transferred to the revision generation task through selected important LLM layers. To measure the layer alignment between the two tasks, we define the layer alignment ratio $r$:
\begin{equation}
    r=\frac{|\mathcal{S}_{pre} \cap \mathcal{S}_{gen}|}{|\mathcal{S}_{gen}|},
    \label{eq:ratio}
\end{equation}
which measures the percentage of the shared LLM layers across the two tasks. 

\subsection{Intention-Adaptive LLM Fine-Tuning}
Given that text revisions are commonly motivated by the writer's diverse intentions, we develop an intention-adaptive multi-task PEFT framework to fine-tune LLMs on the tasks of intention prediction and revision generation. Note that we fine-tune the two tasks sequentially, rather than simultaneously, to alleviate computational burdens. The two tasks share weights across LLM layers and maintain separate prediction and generation headers.

\paragraph{Single-Intent Revision Objective} Given a pair of revisions $\{ \mathcal{X},\mathcal{Y}\}$ and intention labels $\mathcal{I}=\{\mathcal{I}_1,\mathcal{I}_2,\dots,\mathcal{I}_n\}$, an intention prediction header aims to learn a label distribution through a softmax function and a cross-entropy loss $\mathcal{L}_{pre}$:
\begin{equation}
    \mathcal{P}(\hat{\mathcal{I}}_i)=\frac{\exp \left(\mathcal{\hat{\mathcal{I}}}_i\right)}{\sum_{i=1}^n \exp \left(\mathcal{\hat{\mathcal{I}}}_i\right)},
\end{equation}
\vspace{-.8em}
\begin{equation}
    \mathcal{L}_{pre} = - \sum_{i=1}^{n} \mathcal{I}_i \log(\hat{\mathcal{I}}_i),
\end{equation}
\noindent where $\hat{\mathcal{I}}_i$ is a predicted intention. In contrast, a revision generation header aims to generate revised text $\hat{\mathcal{Y}}$, given the original text $\mathcal{X}$ and revision intention $\mathcal{I}$. Although the generation task also uses cross-entropy loss $\mathcal{L}_{gen}$, it is optimized to maximize the probability of the next token $y_t$ rather than the intention label:
\begin{equation}
\mathcal{P}(\hat{\mathcal{Y}} \mid \mathcal{X}, \mathcal{I}) = \prod_{t=1}^{d} \mathcal{P}(y_t \mid \mathcal{X}, \mathcal{I}, y_{<t}),
\end{equation}
\vspace{-.8em}
\begin{equation}
   \mathcal{L}_{gen} = -\sum_{t=1}^d \log \mathcal{P}\left(y_t \mid \mathcal{X}, \mathcal{I}, y_{<t}\right),
\end{equation}
\noindent where $y_{<t}$ are the tokens before $y_t$ and $d$ is the number of tokens in the sequence. 

\paragraph{Multi-Intent Revision Objective} In the cases that text revisions are driven by multiple intentions, the revised sequences need to simultaneously fulfill all the revision purposes $\mathcal{I}=\{\mathcal{I}_1,\mathcal{I}_2,\dots,\mathcal{I}_n\}$. Thus, an intention prediction header aims to learn multiple labels in $\mathcal{I}$. Here, a multi-label cross-entropy loss $\mathcal{L}_{pre}^{\prime}$ is formulated as:
\begin{equation}
    \mathcal{L}_{pre}^{\prime} = - \sum_{i}^{n} \left[ \mathcal{I}_i \log(\hat{\mathcal{I}}_i) + (1 - \mathcal{I}_i) \log(1 - \hat{\mathcal{I}}_i) \right],
\end{equation}
\noindent where the predicted label $\hat{\mathcal{I}_i}$ is passed to a sigmoid function for $\hat{\mathcal{I}}_i = \sigma(z_i)=\frac{1}{1+e^{-z_i}}$. Similarly, the optimized probability of the next tokens in the revised text is formulated as:
\vspace{-1.5em}
\begin{align}
\mathcal{P^{\prime}}(\mathcal{\hat{Y}} \mid \mathcal{X}, \mathcal{I}_1, \mathcal{I}_2, \ldots, \mathcal{I}_k) 
&= \prod_{t=1}^{d} \mathcal{P}(y_t \mid \mathcal{X}, \nonumber \\
& \mathcal{I}_1, \mathcal{I}_2, \ldots, \mathcal{I}_n, y_{<t}),
\end{align}
\vspace{-2.2em}
\begin{equation}
   \mathcal{L}_{gen}^{\prime} = -\sum_{t=1}^d \log \mathcal{P}\left(y_t \mid \mathcal{X}, \mathcal{I}_1, \mathcal{I}_2, ..., \mathcal{I}_n,y_{<t}\right),
\end{equation}

\noindent where $d$ is the number of tokens in the sequence. 

\begin{algorithm}[tb]
\caption{Intention-adaptive layer-wise PEFT}
\label{algo:mtl_shared}
\textbf{Output}: An optimized LLM $\mathcal{M}^{\prime}$ \\
\textbf{Input}: Datasets $\mathcal{D}_{pre}$, $\mathcal{D}_{gen}$ for Task $\mathcal{T}_{pre}$ and $\mathcal{T}_{gen}$, \\
Fine-tuning steps $\mathcal{K}$, LLM $\mathcal{M}$ with PEFT adapters \\
\textbf{Parameter}: Important layer $\mathcal{S}=\emptyset$, \\
redundant layer $\mathcal{\bar{S}}=\emptyset$
\begin{algorithmic}[1]
    \FOR{$k = 1$ to $\mathcal{K}$}
        \STATE Compute fine-tuning loss $\mathcal{L}_{\text{$pre$}}$ and layer-wise gradient norm $\mathcal{N}_{k}$ on $\mathcal{T}_{pre}$
        \STATE Use $\mathcal{N}_{k}$ to obtain important and redundant layer sets $\mathcal{S}_k$, $\mathcal{\bar{S}}_k$ by Eq.~\ref{eq:likelihood},~\ref{eq:variance}
        \STATE Activate LLM layer $m \in \mathcal{S}_k$
        \STATE freeze LLM layer $m^{\prime} \in \mathcal{\bar{S}}_k$
    \ENDFOR
    \STATE Compute layer-wise importance frequency $\mathcal{G}$ by Eq.~\ref{eq:frequency}  and Task $\mathcal{T}_{pre}$
    \STATE Use $\mathcal{G}$ to obtain another important and redundant layer sets $\mathcal{S}$, $\mathcal{\bar{S}}$ by Eq.~\ref{eq:likelihood},~\ref{eq:variance}
    \FOR{$k = 1$ to $\mathcal{K}$}
        \STATE Activate LLM layer $m \in \mathcal{S}$
        \STATE Freeze LLM layer $m^{\prime} \in \mathcal{\bar{S}}$
        \STATE Compute fine-tuning loss $\mathcal{L}_{\text{$gen$}}$ on $\mathcal{D}_{gen}$
    \ENDFOR
\end{algorithmic}
\end{algorithm}

\subsection{Layer Update with PEFT}
While standard PEFT methods have reduced computational overhead, their efficiency can be further enhanced by selectively fine-tuning a subset of layers. In particular, we integrate Low-Rank Adaptation (LoRA)~\cite{hu2022lora} into important LLM layers, i.e., important LoRA layers, for efficient low-resource tuning, while freezing those of redundant layers. As shown in Figure~\ref{fig:framework}, both intention prediction and revision generation tasks are fine-tuned based on shared adapter-based PEFT (LoRA). The pseudo-algorithm is described in Algorithm~\ref{algo:mtl_shared}. 

\section{Corpora}
\label{sec:corpora}

\paragraph{ArgRevision}Text revision is rarely annotated because its annotation is a costly process; thus, we use a previously collected and annotated argument revision corpus named ArgRevision~\cite{liu-litman-2025-ir-tuning}, which contains 660 pairs of essay drafts written by students in grades four to eight. The ArgRevision corpus includes three essay drafts, where students write initial drafts, and revise their essays in the second drafts based on provided feedback, and further revise their essays in the third drafts based on another round of feedback. The pairs of drafts (e.g., draft1-draft2 and draft2-draft3) are annotated with sentence-level revisions, using their dedicated intention labels: \textit{relevant, irrelevant, repeated, linked claim-evidence (LCE), not LCE, commentary} and \textit{others}. Detailed label taxonomy and examples are described in Appendix~\ref{sec:argrevision_example}. The intention label statistics are shown in Table \ref{table:sentence-stats}.

\paragraph{ITERATER} We utilize a publicly available revision corpus named ITERATER~\cite{du-etal-2022-understanding-iterative}, which annotates 4,018 sentence-level text revisions from Wikipedia, ArXiv, and news articles.  
The corpus contains six intention labels: \textit{clarity}, \textit{fluency}, \textit{coherence}, \textit{style}, \textit{meaning-changed}, and \textit{others}. 
Detailed label taxonomy and annotation examples are provided in Appendix~\ref{sec:iterater_example}. The label statistics are shown in Table~\ref{table:iterator-corpus-stats}.

\begin{table}[t]\small
\centering
\begin{adjustbox}{width=1\columnwidth}
\begin{tabular}{c|ccccccc|c}
\toprule
  Intention                & Relevant & Irrelevant & Repeated  & LCE    & not LCE  & Commentary & Others &  Total                      \\ \midrule
Add               & 1,774    & 397        & 225         & 1,069  & 317      & 425        & 435    & 4,642                  \\
Delete            & 598      & 102        & 45          & 318    & 121      & 194        & 126     & 1,504                  \\
Modify            & 138      & 26         & 6           & 105    & 23       & 41         & 108     & 447                    \\ \midrule
Total             & 2,510    & 525        & 276         & 1,492  & 461      & 660        & 669    & 6,593                  \\ \bottomrule
\end{tabular}
\end{adjustbox}
\vspace{-.1in}
\caption{\label{table:sentence-stats}
Statistics of revision intentions in ArgRevision.
}
\vspace{-0.1in}
\end{table}

\begin{table}[tb]\small
\centering
\begin{adjustbox}{width=0.88\columnwidth}
\begin{tabular}{c|cccccc|c}
\toprule
   Intention    & Clarity & Fluency & Coherence & Style & Meaning & Others & Total \\ \midrule
Add    & 94      & 208     & 37        & 2     & 342             & 6      & 689   \\
Delete & 282     & 111     & 180       & 20    & 11              & 11     & 615   \\
Modify & 1,225    & 623     & 176       & 106   & 543             & 41     & 2,714  \\ \midrule
Total  & 1,601    & 942     & 393       & 128   & 896             & 58     & 4,018  \\ \bottomrule
\end{tabular}
\end{adjustbox}
\vspace{-.1in}
\caption{Statistics of revision intention in ITERATER.}
\label{table:iterator-corpus-stats}
\vspace{-.2in}
\end{table}

\paragraph{Data Preprocessing}
We preprocess the two corpora for sentence-level and document-level revision generation. In ArgRevision, the original sentence is empty in \textit{adding}, the revised sentence is empty in \textit{deleting}; only \textit{modifying} revisions are used, in which, however, only 447 examples are collected (see Table~\ref{table:sentence-stats}). Hence, we excluded ArgRevision from sentence-level revision generation; instead, we used it for document-level generation, where a document is revised based on multiple intentions. To simplify the task, we added an <edit> tag before and after an edited sentence to note a text revision in the ArgRevision dataset. Regarding the ITERATER corpus, we utilize its sentence-level dataset, ITERATER-sent, which is single-intent revisions, and its document-level dataset, ITERATER-doc, which contains multi-intent revisions (see Figure~\ref{fig:intro-example}). We exclude~\textit{others} label in both corpora, yielding six intentions in ArgRevision and five intentions in ITERATER. The data split for the three datasets is summarized in Table~\ref{table:corpus-split}. The following instruction prompts are used for the intention prediction and revision generation tasks, respectively:     

\begin{table}[tb]\small
\centering
\begin{adjustbox}{width=0.85\columnwidth}
\begin{tabular}{c|ccc}
\toprule
  Corpus    & ITERATER-sent & ITERATER-doc & ArgRevision \\ \midrule
Train/Val/Test & 3,215 / 385 / 360 & 481 / 27 / 51 & 528 / 66 / 66        \\ 
\bottomrule
\end{tabular}
\end{adjustbox}
\vspace{-.1in}
\caption{Data splits across different corpora.}
\label{table:corpus-split}
\vspace{-1.8em}
\end{table}

\begin{itemize}[leftmargin=*,itemsep=-5pt, topsep=0pt]
    \item \noindent {\textbf{Intention Prediction}: Identify the intention of the revision between the original text and the revised text. The possible intentions include}: $\mathcal{I}$.  {Original Text}: {$\mathcal{X}$}.  {Revised Text}: {$\mathcal{Y}$}.
    \item \noindent  {\textbf{Revision Generation}: Revise the original text based on the intention $\mathcal{I}$}. {Original Text}: {$\mathcal{X}$}. {Revised Text}: $\mathcal{Y}$.
\end{itemize}

\begin{figure*}[tb]
    \centering
    \includegraphics[width=2.09\columnwidth]{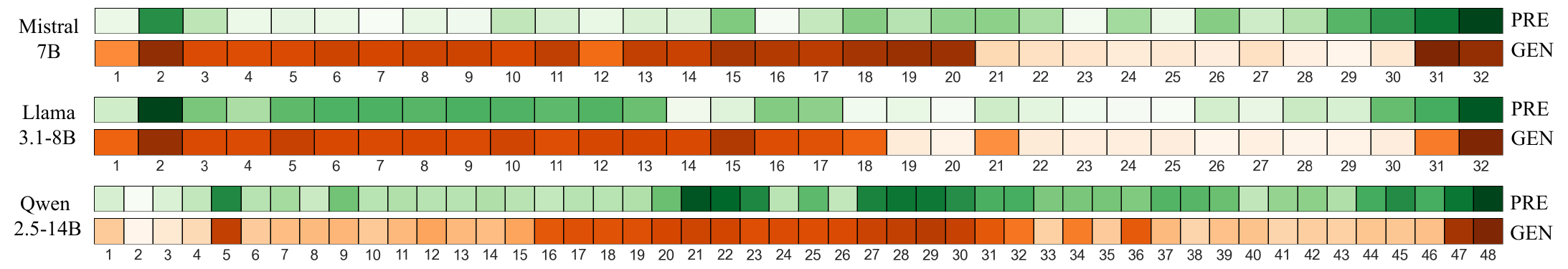}
    \vspace{-1.99em}
    \caption{The layer-wise importance score alignment between the intention prediction (green boxes) and the revision generation (brown boxes) tasks while fine-tuning LLMs on the ITERATER-sent. All PEFT uses LoRA. The dark colors indicate high scores (important layers) while the light colors indicate low scores (redundant layers).}
    \vspace{-.9em}
    \label{fig:layer-gradient}
\end{figure*}

\section{Experiments}
We use cutting-edge LLMs, i.e., upgraded Mistral-7B-Instruct-v0.3 (Mistral-7B) known for efficient inference~\cite{jiang2023mistral7b}, Llama3.1-8B-Instruct (Llama3.1-8B) good for general NLP tasks~\cite{grattafiori2024llama3herdmodels}, and Qwen2.5-14B with relatively larger parameters and strong multilingual capability~\cite{qwen2.5}. We compare our Intention-Tuning to the following baselines:

\begin{itemize}[leftmargin=*,itemsep=-5pt, topsep=0pt]
    \item \textbf{Copy-Baseline}: We copy the original text as revised text for a non-edit baseline.
    \item \textbf{BART-Baseline}: We train  BART-large ~\cite{lewis2019bart} as a small language model baseline.
    \item \textbf{ICL-Baseline}: We use in-context learning as a non-finetuned LLM baseline. The used prompt is based on~\citet{raheja-etal-2023-coedit} and~\citet{ziegenbein-etal-2024-llm}, as described in Appendix~\ref{sec:in-context-learning}.
    \item \textbf{CoT-Baseline}: We use in-context learning with a chain-of-though prompt~\cite{wei-2022-chain-of-thought} as another non-finetuned baseline in Appendix~\ref{sec:in-context-learning}.
    \item \textbf{LISA-Baseline}: We fine-tune randomly selected four layers based on~\citet{pan2024lisa}. We apply PEFT (LoRA) to the LLM layers.
    \item \textbf{IST-Baseline}: We compute layer-wise importance scores based on~\citet{yao-etal-2024-layer} and sample the top eight LLM layers for PEFT (LoRA).
    \item \textbf{IR-Baseline}: We use gradient norms based on~\citet{liu-litman-2025-ir-tuning} to dynamically select a subset of LLM layers for PEFT (LoRA).
    \item \textbf{Full-Finetuning}: We fine-tune all LLM layers with PEFT (LoRA) as a strong baseline.
\end{itemize}

In the implementation, we build the framework using PyTorch and HuggingFace, and optimize task losses using the Adam optimizer on a Nvidia A100 GPU. We set the batch size and max length as 16 and 256 for ITERATER-sent, four and 1,024 for ITERATER-doc, two and 1,536 for ArgRevision, respectively. We use a learning rate of 2e-4, set the maximum epochs to two, select important and redundant layers at every step, and log the loss every 20 steps. We fine-tune LLMs on the training sets, tune hyperparameters on the validation sets, and report \textbf{SARI}~\cite{xu2016optimizing}, \textbf{GLEU}~\cite{napoles-etal-2015-ground}, and Update-ROUGE (\textbf{Update-R})~\cite{iv-etal-2022-fruit} scores on the test sets. The SARI is designed to evaluate addition, deletion, and modification in n-grams; GLEU is customized to penalize changed n-grams; Update-R focuses on updated sentences rather than full text~\cite{shu2024rewritelm}. Detailed metric implementation and hyperparameter settings during fine-tuning, generation, and evaluation are described in Appendix~\ref{sec:hyperparamter}.

\begin{table*}[t]\small
\centering
\begin{adjustbox}{width=1\linewidth}
\begin{tabular}{cc|cccc|cccc|cccc}
\toprule
\multirow{2}{*}{Models} & \multirow{2}{*}{Methods} & \multicolumn{4}{c|}{ITERATER-sent} & \multicolumn{4}{c|}{ITERATER-doc} & \multicolumn{4}{c}{ArgRevision} \\ \cmidrule{3-14} 
 &  & SARI & GLEU & Update-R & Average & SARI & GLEU & Update-R & Average & SARI & GLEU & Update-R & Average \\ \midrule
Copy-Baseline & N/A & 29.46 & 68.86 & 2.50 & 33.61 & 27.17 & 53.49 & 0.00 & 26.89 & 30.07 & 80.60 & 3.75 & 38.14 \\
BART-Baseline & N/A & 31.74 & 66.85 & 35.57 & 44.72 & 41.68 & 47.57 & 48.22 & 45.82 & 16.78 & 2.34 & 15.31 & 11.48 \\ \midrule
\multirow{8}{*}{Mistral-7B} & Full-Finetuning & 37.29 & 64.26 & 56.88 & 52.81 & 42.85 & 53.71 & 32.86 & 43.14 & 36.00 & 62.48 & 14.19 & 37.56 \\ \cmidrule{2-14} 
& ICL-Baseline & 31.43 & 13.10 & \,\,\,61.99* & 35.51 & 37.93 & 23.90 & \textbf{\,\,\,53.69*} & 38.51 & 28.13 & 26.97 & \,\,\,18.79* & 24.63 \\
& CoT-Baseline & 31.53 & 12.48 & \,\,\,61.83* & 35.28 & 36.74 & 20.42 & \,\,\,52.45* & 36.54 & 24.36 & 11.24 & \textbf{\,\,\,22.07*} & 19.22 \\ 
 & LISA-Baseline & 36.68 & \textbf{\,\,\,66.33*} & 56.71 & \,\,\,53.24* & 35.86 & \,\,\,53.91* & \,\,\,35.66* & 41.81 & 34.14 & \textbf{62.23} & 11.28 & 35.88 \\
 & IST-Baseline & \textbf{\,\,\,39.42*} & 64.04 & \textbf{\,\,\,67.95*} & \textbf{\,\,\,57.14*} & 38.03 & \,\,\,54.53* & 30.90 & 41.15 & 34.29 & 62.06 & 9.80 & 35.38 \\
 & IR-Baseline & \,\,\,38.60* & 63.79 & \,\,\,59.51* & \,\,\,53.97* & \,\,\,45.03* & \textbf{\,\,\,54.81*} & \,\,\,34.21* & \,\,\,44.68* & \,\,\,36.55* & 61.77 & 13.58 & \textbf{37.30} \\
 & Intention-Tuning & \,\,\,39.33* & 63.72 & \,\,\,65.71* & \,\,\,56.25* & \textbf{\,\,\,45.40*} & \,\,\,54.58* & \,\,\,35.15* & \textbf{\,\,\,45.04*} & \textbf{\,\,\,36.78*} & 62.07 & 12.79 & 37.21 \\ \midrule
\multirow{8}{*}{Llama3.1-8B} & Full-Finetuning & 42.52 & 63.64 & 77.26 & 61.14 & 47.61 & 54.79 & 51.76 & 51.39 & 37.80 & 62.79 & 15.63 & 38.74 \\ \cmidrule{2-14} 
& ICL-Baseline & 38.96 & 27.93 & 69.12 & 45.34 & 39.03 & 29.58 & \textbf{\,\,\,54.01*} & 40.87 & 35.66 & 52.24 & \textbf{\,\,\,19.08*} & 35.66 \\
& CoT-Baseline & 37.77 & 25.64 & 66.75 & 43.39 & 39.24 & 18.45 & 47.66 & 35.12 & 33.09 & 48.42 & 15.54 & 32.35 \\
 & LISA-Baseline & 36.86 & 62.63 & 74.63 & 58.04 & 38.66 & 54.03 & 46.11 & 46.27 & 34.38 & \textbf{\,\,\,64.43*} & 12.04 & 36.95 \\
 & IST-Baseline & 39.28 & 63.16 & 76.82 & 59.75 & 41.68 & 54.76 & 48.98 & 48.47 & 34.38 & \,\,\,64.38* & 12.20 & 36.99 \\
 & IR-Baseline & 41.77 & \,\,\,63.88* & 76.73 &60.79 & 46.94 & 54.74 & 48.79 & 50.16 & 37.19 & 62.48 & \,\,\,16.83* & \,\,\,38.83* \\
 & Intention-Tuning & \textbf{42.10} & \textbf{\,\,\,64.08*} & \textbf{\,\,\,77.99*} & \textbf{\,\,\,61.39*} & \textbf{\,\,\,48.77*} & \textbf{\,\,\,54.86*} & {\,\,\,52.08*} & \textbf{\,\,\,51.90*} & \textbf{37.37} & \,\,\,63.72* & \,\,\,16.58* & \textbf{\,\,\,39.22*} \\ \midrule
\multirow{8}{*}{Qwen2.5-14B} & Full-Finetuning & 41.95 & 60.89 & 76.08 & 59.64 & 50.13 & 55.25 & 38.60 & 47.99 & 41.38 & 60.96 & 25.35 & 42.56 \\ \cmidrule{2-14} 
& ICL-Baseline & 34.00 & 29.80 & 61.61 & 41.80 & 38.28 & 27.86 & \,\,\,51.85* & 39.33 & 33.82 & 55.90 & 15.02 & 34.91 \\
& CoT-Baseline & 35.72 & 18.52 & 65.34 & 39.86 & 38.09 & 19.00 & \textbf{\,\,\,52.63*} & 36.57 & 33.07 & 50.75 & 17.81 & 33.88 \\
 & LISA-Baseline & 39.82 & \textbf{\,\,\,63.13*} & 74.44 & 59.13 & 37.63 & 53.89 & 32.76 & 41.43 & 34.14 & {\,\,\,63.65*} & 13.40 & 37.06 \\
 & IST-Baseline & 41.12 & \,\,\,63.09* & 75.71 & \,\,\,59.97* & 41.06 & 53.84 & \,\,\,38.93* & 44.61 & 36.22 & \textbf{\,\,\,63.81*} & 14.29 & 38.11 \\
 & IR-Baseline & \textbf{41.79} & 60.60 & \,\,\,76.88* & \,\,\,59.76* & 45.07 & 51.80 & \,\,\,39.51* & 45.46 & 37.51 & 60.91 & \textbf{21.74} & 40.05 \\
 & Intention-Tuning & 41.13 & \,\,\,62.94* & \textbf{\,\,\,77.50*} & \textbf{\,\,\,60.52*} & \textbf{46.14} & \textbf{54.53} & \,\,\,40.67* & \textbf{47.11} & \textbf{38.08} & \,\,\,63.09* & 19.94 & \textbf{40.37} \\
\bottomrule
\end{tabular}
\end{adjustbox}
\vspace{-.4em}
\caption{The performance of Intention-Tuning and baselines on ITERATER-sent, ITERATER-doc, and ArgRevision datasets. The bold numbers represent the best results, and the asterisks (*) indicate that the results are better than those of Full-Finetuning.
}
\label{tab:peft-comparision}
\vspace{-0.5em}
\end{table*}

\section{Results}
\subsection{LLM Layer Alignment}
\label{sec:layer_alignment}
To investigate layer-wise alignment between the intention prediction and revision generation tasks, Figure~\ref{fig:layer-gradient} visualizes layer-wise importance scores, i.e., average gradient norms throughout fine-tuning, for the two tasks. Regarding Mistral-7B, a few top, middle, and bottom layers in the prediction task (green) have high importance scores, e.g., dark green boxes in layers 2, 17 to 22, and 29 to 32, which are mostly selected for fine-tuning. In contrast, the top and bottom layers in the generation task have high importance scores, e.g., dark brown boxes in layers 1 to 20 and 31 to 32, which suggests that these layers are frequently selected during revision generation. Also, the layers in the lower middle have low importance scores, e.g., light brown boxes in layers 21 to 30, which suggests these layers are regarded as redundant and are mostly frozen. In terms of Llama3.1-8B, the top to middle and a few bottom layers are both identified as important layers in the two tasks. For Qwen2.5-14B, the middle and bottom layers are mainly used for the prediction task, while similar middle and bottom layers with slightly shifted locations are mostly used for the generation task. Additionally, several important and redundant layers are located in consecutive positions, suggesting that neighboring layers may share similar fine-tuning patterns. 

Despite differences across three LLMs, the important and redundant layers are mostly aligned between the prediction and generation tasks, e.g., dark green and dark brown boxes are mostly aligned\footnote{Although the colored boxes indicate the LLM layer-wise importance scores, they do not exactly reflect the binary important and redundant layers as well as their alignment across tasks, since there is a threshold to split layer-wise importance scores. Their layer alignment ratio is shown in Table~\ref{tab:importance-score-correlation}.}, and particularly aligned for layers 2 and 32 in Mistral-7B and Llama3.1-8B, as well as layer 5 in Qwen2.5-14B. Similar patterns are observed in Figure~\ref{fig:layer-gradient-iterater-doc} and Figure~\ref{fig:layer-gradient-argrevision} in the Appendix regarding the ITERATER-doc and ArgRevision datasets. These findings reveal that despite the unique designs across different LLMs, their layers exhibit similar patterns. This consistency demonstrates the feasibility of aligning the intention prediction and revision generation tasks through selected LLM layers for Intention-Tuning, which answers \textbf{RQ1}.

\subsection{Performance Comparison}
\label{sec:performance_comparision}
We evaluate Intention-Tuning in Table~\ref{tab:peft-comparision}. The Copy-Baseline that copies the original text as revised text has low SARI and Update-R scores, but relatively high GLEU scores. This suggests that the original and revised texts have a high overlap; SARI and Updated-R are key metrics for evaluating revision quality. Also, the BART-Baseline exhibits low performance on ITERATER and poor results on ArgRevision due to its limited capability to generate complex revisions (e.g., argument essay revisions), which implies the need for LLM solutions. In ITERATER-sent, Intention-Tuning achieves the best results on Llama3.1-8B, and the best Update-R and Average scores on Qwen2.5-14B. Although not competitive regarding Mistral-7B, Intention-Tuning achieves close performance to the best SARI score, and generally outperforms Full-Finetuning. In ITERATER-doc, Intention-Tuning largely outperforms the baselines, and sometimes better than Full-Finetuning, regarding Llama3.1-8B and Qwen2.5-14B. Although ICL and CoT baselines achieve relatively high Update-R scores, they exhibit low GLEU scores, indicating that while in-context learning produces revision-like outputs, the edits do not align well with the reference revisions. In ArgRevision, while Intention-Tuning mostly performs better on average, the key metrics are lower than those in the ITERATER datasets, which again suggests that the argument revisions are more complicated than the scientific and news revisions. In general, Llama3.1-8B and Qwen2.5-14B with Intention-Tuning achieve better Average scores than the IR-Baseline across three datasets, which highlights the advantages of transferring layer representations from intention prediction to revision generation.

We argue that the degree of the LLM layer alignment between the two revision tasks could impact its performance, given that both Llama3.1-8B and Qwen2.5-14B exhibit higher layer-wise alignment and higher performance, compared to Mistral-7B, as shown in Figure~\ref{fig:layer-gradient} and Table~\ref{tab:peft-comparision}. In particular, we employ the layer alignment ratio metric, as defined in Equation~\ref{eq:ratio}, to measure the layer-wise alignment between the two tasks.  Table~\ref{tab:importance-score-correlation} shows that more than half of the layers could be aligned using Intention-Tuning, of which Llama3.1-8B generally has high alignment, and Mistral-7B performs better in ITERATER-doc than the other two datasets. These observations are in line with the performance in Table~\ref{tab:peft-comparision}. Hence, we reason that Intention-Tuning would work best for task-agnostic LLM layers where the layer-wise performance remains consistent across tasks (e.g., layers 2 to 13, 16, 17, 31, and 32 in Llama3.1-8B, as shown in Figure~\ref{fig:layer-gradient}), which will be studied in future work.  

\begin{table}[tb]
\centering
\begin{adjustbox}{width=0.8\columnwidth}
\begin{tabular}{c|ccc}
\toprule
LLMs & ITERATER-sent & ITERATER-doc & ArgRevision \\ \midrule
Mistral-7B & 50.00 \% & \textbf{68.18} \% & 66.67 \% \\
Llama3.1-8B & \textbf{83.33} \% & 56.25 \% & \textbf{72.73} \% \\
Qwen2.5-14B & 63.64 \% & 52.38 \% & 56.67 \% \\ \bottomrule
\end{tabular}
\end{adjustbox}
\vspace{-.5em}
\caption{The layer alignment ratio (percentage) between the intention prediction and revision generation tasks.}
\label{tab:importance-score-correlation}
\vspace{-0.1in}
\end{table}

\begin{table}[tb]\small
\centering
\begin{adjustbox}{width=0.94\columnwidth}
\begin{tabular}{c|ccccc}
\toprule
Methods & Clarity & Fluency & Coherence & Style & Meaning \\ \midrule
 Full-Finetuning & 57.25 & 75.78 & 64.39 & 60.36 & 44.15 \\ \midrule 
  ICL-Baseline & 37.96 & 63.55 & 44.93 & 26.20 & 35.69 \\
 CoT-Baseline & 35.53 & 55.68 & 41.14 & 41.00 & 32.92 \\
 LISA-Baseline & 55.68 & 69.43 & \textbf{\,\,\,65.05*} & 46.81 & 40.88 \\
  IST-Baseline & 56.53 & 73.29 & 63.29 & \textbf{55.74} & 43.79 \\
  IR-Baseline & \textbf{\,\,\,58.69*}   & 74.48 & 60.99 & 54.02 & 43.23 \\
 Intention-Tuning & \,\,\,58.46* & \textbf{74.72} & \,\,\,64.70* & 54.86 & \textbf{\,\,\,45.52*} \\ 
\bottomrule
\end{tabular}
\end{adjustbox}
\vspace{-0.5em}
\caption{The average scores on ITERATER-sent for the Llama3.1-8B model. The bold numbers represent the best results, and the asterisks indicate that the results are better than Full-Finetuning.}
\label{tab:sari-comparision}
\vspace{-.18in}
\end{table}

We evaluate the contribution of individual intention to the revision generation task on ITERATER-sent. Table~\ref{tab:sari-comparision} shows Intention-Tuning achieves the best on \textit{fluency} and \textit{meaning-changed}, and close to IR-Baseline on \textit{clarity}, but lower than the PEFT baselines on \textit{coherence} and \textit{style}. This might be because the two intentions have the smallest number of annotations used for the fine-tuning, as shown in Table~\ref{table:iterator-corpus-stats}. Although lower than the PEFT results, Llama3.1-8B outperforms Full-Finetuning on \textit{clarity}, \textit{coherence}, and \textit{meaning-changed}, as well as ICL and CoT on all intentions. These observations suggest Intention-Tuning is effective across various revision corpora, which answers \textbf{RQ2}.

\begin{figure}[t]
    \centering
    \vspace{-.1in}
    \includegraphics[width=1.0\columnwidth]    {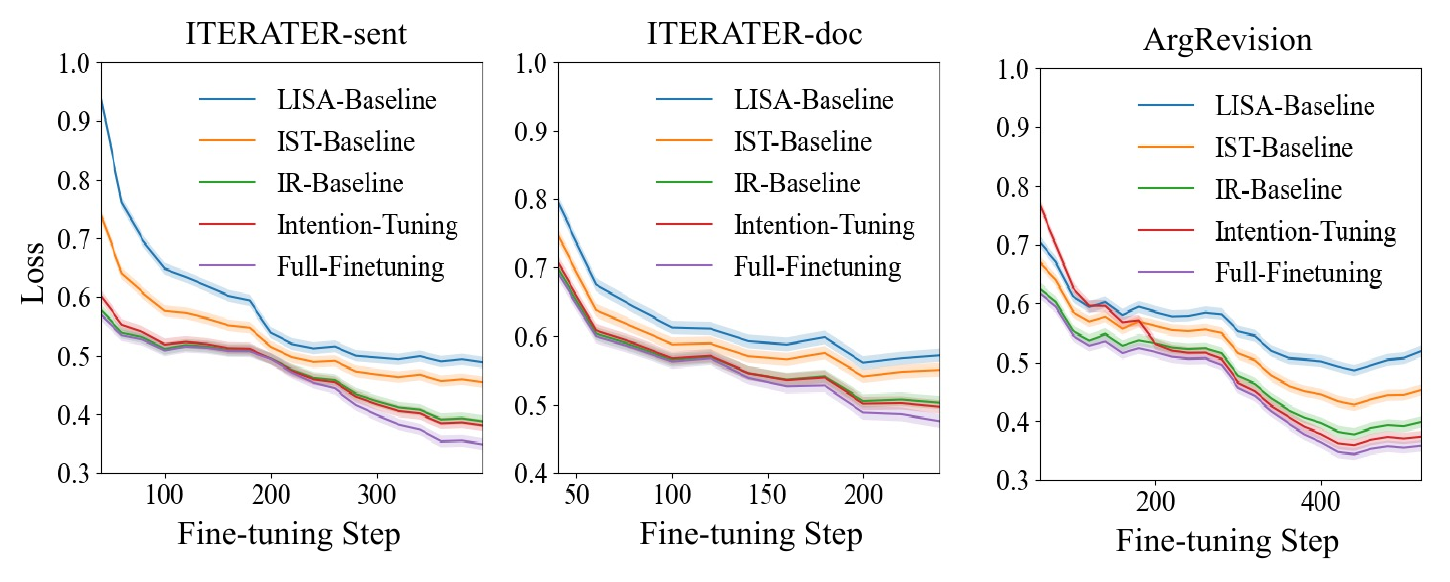}
    \vspace{-.34in}
    \caption{Llama3.1-8B fine-tuning loss on the training set of the datasets. All PEFT uses LoRA.}
    \label{fig:fine-tuning-loss}
    \vspace{-.21in}
\end{figure}

\subsection{Efficiency Evaluation} 
To evaluate Intention-Tuning efficiency, we visualized the loss convergence of Llama3.1-8B, as it generally exhibits high performance in Table~\ref{tab:peft-comparision} and high layer-wise alignment in Table~\ref{tab:importance-score-correlation}. Figure~\ref{fig:fine-tuning-loss} shows that Intention-Tuning with LoRA consistently achieves faster convergence than the other baselines and is even close to Full-Finetuning. Similar patterns are observed while using other PEFT adapters, i.e., the recent DoRA~\cite{pmlr-v235-liu24bn} and classic Bottleneck~\cite{houlsby2019parameter}, as described in Appendix~\ref{sec:convergence_adapter}. These results demonstrate that Intention-Tuning remains efficient during fine-tuning. Additionally, we measured peak GPU memory allocations while fine-tuning Llama3.1-8B using standard PEFT versus Intention-Tuning with a batch size of one. Table~\ref{tab:fine-tuning-memory} shows that Intention-Tuning uses less GPU memory than standard PEFT methods across all adapters and three datasets. Among the methods, Intention-Tuning generally saves 5\% to 14\%. We further compare Intention-Tuning performance using different adapters in Table~\ref{tab:adapter-comparision}. The results suggest that LoRA stands out for its minimal memory allocation and best performance on the ITERATER datasets. While Bottleneck achieves the highest performance on ArgRevision, it requires 10\% more resources than LoRA, and LoRA’s performance remains competitive. These findings indicate that Intention-Tuning is efficient in both fine-tuning convergence and GPU memory allocations, and the findings generalize across different PEFT adapters and datasets, which answers \textbf{RQ3}.

\begin{table}[tb]\small
\centering
\begin{adjustbox}{width=0.99\columnwidth}
\begin{tabular}{c|cc|c|cc|c|cc|c}
\toprule
\multirow{2}{*}{Adapter} & \multicolumn{3}{c|}{ITERATER-sent} & \multicolumn{3}{c|}{ITERATER-doc} & \multicolumn{3}{c}{ArgRevision} \\ \cmidrule{2-10} 
 & Std & IntT & Save & Std & IntT & Save & Std & IntT & Save \\ \midrule
LoRA & 19 & 18 & 5\% & 26 & 24 & 8\% & 30 & 28 & 7\% \\
DoRA & 20 & 19 & 5\% & 31 & 25 & 19\% & 36 & 31 & 14\% \\
Bottleneck & 23 & 20 & 13\% & 30 & 28 & 7\% & 33 & 31 & 6\% \\
\bottomrule
\end{tabular}
\end{adjustbox}
\vspace{-.5em}
\caption{Llama3.1-8B GPU allocations (Gigabytes) between standard (Std) PEFT and Intention-Tuning (IntT).}
    \label{tab:fine-tuning-memory}
    \vspace{-.05in}
\end{table}

\begin{table}[tb]
\centering
\begin{adjustbox}{width=0.89\columnwidth}
\begin{tabular}{c|ccc}
\toprule
Adapter & ITERATER-sent & ITERATER-doc & ArgRevision \\ \midrule
LoRA & \textbf{61.39} & \textbf{51.90} & 39.22 \\
DoRA & 59.95 & 51.28 & 37.97 \\
Bottleneck & 59.58 & 49.83 & \textbf{39.42} \\ \bottomrule
\end{tabular}
\end{adjustbox}
\vspace{-.5em}
\caption{Llama3.1-8B with Intention-Tuning. The results are the average scores of the metrics.}
\label{tab:adapter-comparision}
\vspace{-0.15in}
\end{table}

\section{Qualitative Analysis} 
\begin{figure}[tb!]
    \centering
    \includegraphics[width=1\linewidth]{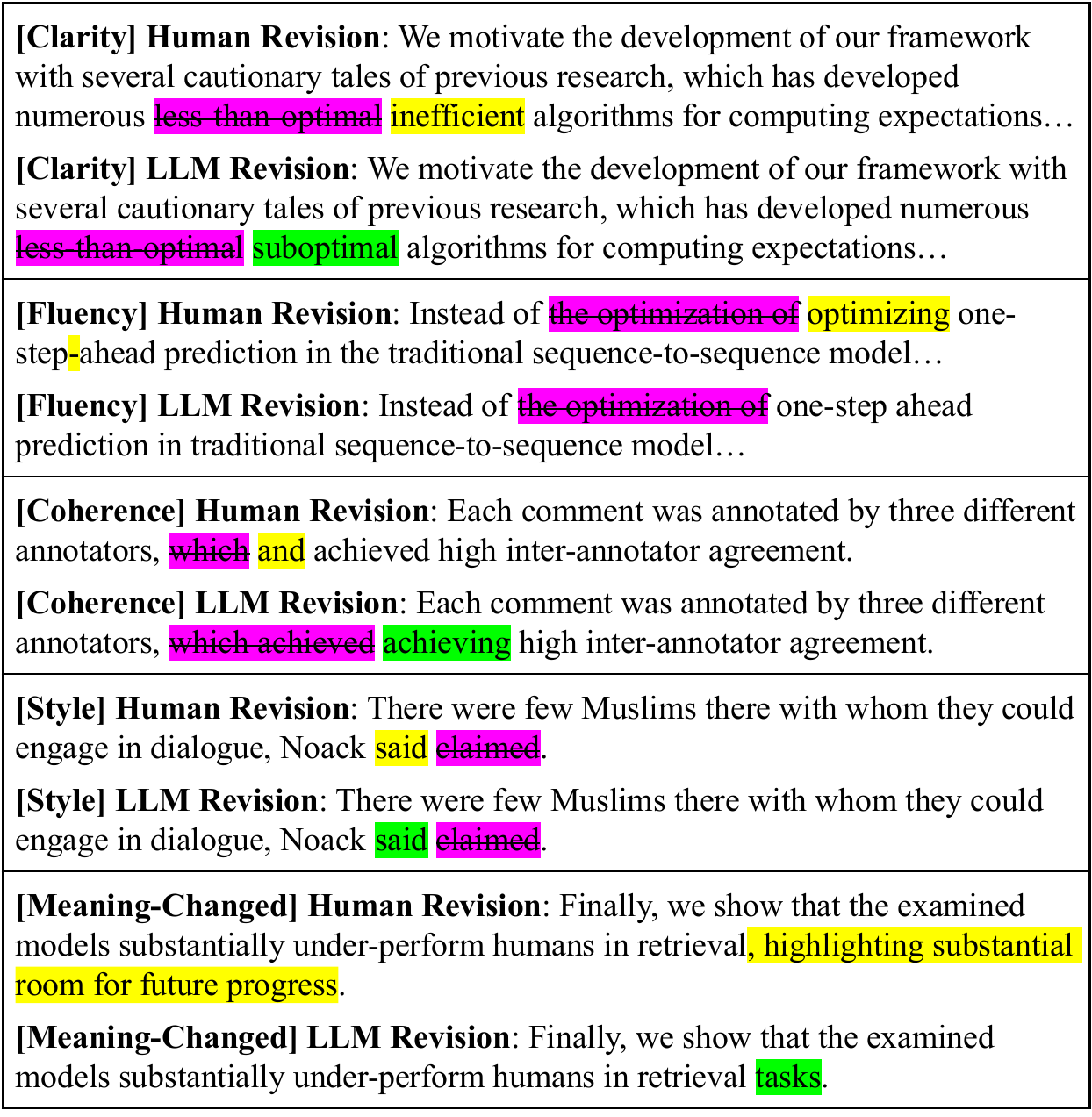}
    \vspace{-.2in}
    \caption{The example of the human and generated revisions by Llama3.1-8B with Intention-Tuning on the ITERATER-sent dataset. The yellow and green colors denote additions, and the purple denotes deletions.}
    \label{fig:example-human-llm}
    \vspace{-.1in}
\end{figure}

We show revision examples in Figure~\ref{fig:example-human-llm}, comparing those edited by humans and generated by Llama3.1-8B using Intention-Tuning on the ITERATER-sent corpus. The human revision identifies a \textit{clarity} issue and replaces the phrase ``less than optimal'' with ``inefficient'', to explain computational inefficiency, while the LLM replaces it with ``suboptimal'', which improves the \textit{clarity} differently. Regarding \textit{fluency}, the human emphasizes the ``optimizing'' while the LLM makes a concise edit. In \textit{coherence}, the human uses ``and'' to connect two actions, and the LLM employs the word ``achieving'' to link the outcome of the annotation, which achieves a more cohesive sentence flow. For \textit{style}, those two have exactly the same revisions. In terms of \textit{meaning-changed}, the human adds additional content to draw out the implications of the finding; however, the LLM only makes a superficial change, which is less informative. Although LLM revisions do not entirely match those of humans, they are typically effective in reflecting the writer’s actual intentions. Regarding multi-intent revisions, LLMs attempt to perform minimal edits, but sometimes generate hallucinated revisions that do not accurately reflect the writer's intentions. Such examples (e.g., Figure~\ref{fig:example-human-llm-iterater-doc} and~\ref{fig:example-human-llm-erevise} in the Appendix) and their error analysis are described in Appendix~\ref{sec:error_analysis}.
These case studies suggest the task difficulty and potential room for future improvement.

\section{Conclusion} 
We present Intention-Tuning, an intention-adaptive layer-wise LLM fine-tuning framework, which transfers the learned representations from the intention prediction to the revision generation through LLM layers. Experiments suggest that it achieves strong performance across LLMs and diverse corpora, while maintaining fast convergence and low GPU allocation. These results reveal its potential and generalizability for complex revision tasks.

\section*{Ethics Statement}

We utilize existing human-annotated revision corpora, which are collected and annotated for academic purposes, and do not pose any ethical implications, such as those related to identity, gender, and race. The corpora include text revisions from both skilled and less skilled writers, which, instead, increases the diversity of the revision research. Furthermore, we acknowledge the existing issues with LLM-generated revisions, such as inaccuracies and hallucinations, which need to be addressed in future work. Lastly, our work on text revision, particularly for argumentative essay revisions, is critically needed for the development of automated student writing evaluation, which would benefit both the education and NLP communities.

\section*{Limitations}

\paragraph{Hypothesis} Our method is grounded in a well-established revision theory that text revisions are primarily motivated by specific intentions. Specifically, we design a two-step method to model the human revision process, in which writers formulate intentions to revise in one step (intention prediction) and implement intention-based revisions in another (revision generation). Also, based on the theory, the revision process is one-directional from intention to generation, but not the opposite. In other words, optimizing the two tasks simultaneously is not theoretically justifiable in the text revision domain. However, our framework could be extended to support both tasks mutually. 

\paragraph{Algorithm} Our algorithm performs multi-task transfer learning on the selected LLM layers, hypothesizing that the important layers for revision generation are aligned with those for intention prediction. Although the findings in Section~\ref{sec:performance_comparision} suggest layer-wise alignment could be task-agnostic, there exist task-specific layers that exhibit different patterns in different tasks, and the alignment could vary based on different intentions and LLMs. Thus, further investigation into LLM layer alignments across tasks and intentions would be helpful. Additionally, the algorithm demonstrates the feasibility of LLM fine-tuning with small corpora, which is particularly advantageous for low-resource NLP tasks. The method’s utilization in broader generation tasks will require corpora annotated with specific intentions. Moreover, the algorithm transfers the learned intention representations to the revision generation, which initially results in a high fine-tuning loss and later decreases to normal levels. Nevertheless, it does not introduce significant computational overhead compared to standard PEFT, based on our pilot studies. 

\paragraph{LLMs} Our evaluation is constrained by computational limitations with relatively lightweight models. Although larger LLMs such as Llama3.3-70B provide stronger contextual understanding, they exceed the scope of this study. In addition, our experiments rely on small annotated corpora, which limits the generalizability of the results to diverse real-world revisions. Also, we follow the literature that commonly reports single-run results due to the high expenses of the LLM fine-tuning; we will perform a significance test based on multi-run results in future work. Furthermore, we focus on layer-wise PEFT methods and do not investigate other PEFT baselines, such as prompt-tuning. Thus, further investigation into alternative PEFT methods and instruction designs could achieve more robust task-specific fine-tuning results. In addition, we randomly select four layers for the LISA-Baseline and the top eight layers for the IST-Baseline because these methods do not provide optimal solutions for determining the number of layers to use. Since the selection of sampling LLM layers is underexplored, we use the optimal numbers based on prior work~\cite{liu-litman-2025-ir-tuning} to achieve a compromise between performance and efficiency. 

\paragraph{Metrics} We use the standard text revision generation metrics that focus on the edit-edit alignment rather than text-text alignment. Although embedding similarity is helpful for text-text comparison, it might not reflect the actual revision quality, given most of the text in a document is unchanged, and thus has high embedding similarity. To the best of our knowledge, LLM-as-a-Judge is not commonly used for text revision generation tasks because it requires criteria tailored to each revision issue. It is interesting to explore, but it needs human evaluation on the LLM judges themselves before it can be implemented for assessing the revision quality, which will be our future work.

\section*{Acknowledgments}

This research was supported by National Science
Foundation Award \#2202347 and by a supplement made in response to NSF DCL 24-093. We also acknowledge the National Artificial Intelligence Research Resource (NAIRR) Pilot and Voltage Park for contributing to our research results. The opinions expressed are those of the authors and do not represent the views of the institutes. The authors would like to thank the anonymous reviewers and the Pitt PETAL and NLP groups for their valuable feedback on this work.

\bibliography{custom}
\appendix
\section{Appendix}
\label{sec:appendix}

\subsection{Probing Important Layer}
\label{sec:ir_method}
Given an LLM $\mathcal{M}=\left\{m_i\right\}_{i=1}^{\ell}$ consisting of $\ell$ transformer layer $m$, its layer-wise gradient norms are denoted as $\mathcal{N}=\{a_i\}_{i=1}^{\ell}$. 
The objective is to split $\ell$ layers into an important subset $\mathcal{S}$ and a redundant subset $\bar{\mathcal{S}}$, where $\mathcal{S} = \{m_i \mid a_i > \gamma\}
$, $\bar{\mathcal{S}} = \{m_i \mid a_i \leq \gamma\}$, and $\gamma$ is a threshold to obtain. 
Based on prior work~\cite{Liu_Liu_Xie_Jin_Jia_2023,liu-litman-2025-ir-tuning}, the layer splitting task can be formulated as a distribution divergence problem, arguing that important and redundant layers are drawn from different distributions. Thus, this problem can be solved as
\begin{equation}
    \label{eq:likelihood}
    \mathcal{S}^*=\operatorname{argmax}_\mathcal{N} \log \frac{\text { Likelihood }\left(\mathcal{H}_1 \mid \mathcal{N}\right)}{\text { Likelihood }\left(\mathcal{H}_0\right)},
\end{equation}
\noindent where $\mathcal{S}^*$ contains selected important layers, $\mathcal{H}_0$ is a null hypothesis that the importance scores of layers in $\mathcal{M}$ follows a single distribution, and $\mathcal{H}_1$ is an alternative hypothesis that there exists a subset $\mathcal{S}$ of $\mathcal{M}$, where importance scores of layers in $\mathcal{S}$ follow a different distribution from those of the remaining layers in $\bar{\mathcal{S}}$.~\citet{xie2021statistically} suggest the likelihood of the alternative hypothesis can be optimized by minimizing the variances of the importance scores in $\mathcal{N}_{\mathcal{S}}=\left\{{a_i\mid m_i\in\mathcal{S}}\right\}$ and $\mathcal{N}_{\bar{\mathcal{S}}}=\left\{{a_i \mid m_i\in\bar{\mathcal{S}}}\right\}$. Hence, an optimal threshold $\gamma^*$ is obtained by minimizing the sum of variances $\operatorname{Var}(\mathcal{N}_\mathcal{S})$ and $\operatorname{Var}\left(\mathcal{N}_{\bar{\mathcal{S}}}\right)$ such that
\begin{equation}
       \operatorname{Var}(\mathcal{N}_{\mathcal{S}}) + \operatorname{Var}\left(\mathcal{N}_{\bar{\mathcal{S}}}\right) <= \operatorname{Var}(\mathcal{N}).
       \label{eq:variance}
\end{equation}

\subsection{ArgRevision Examples}
\label{sec:argrevision_example}
We use the argument essays corpus previously described in~\citet{liu-litman-2025-ir-tuning}. The corpus addresses a crucial need for argument revision in automated student writing assessment, which offers valuable contributions to both the education and NLP communities. The annotation taxonomy~\cite{afrin2020RER} includes:

\begin{itemize}
[leftmargin=*,itemsep=-5pt, topsep=2pt]
\item \textbf{Relevant}: relevant revision is about examples or details that are relevant to the claim.
\item \textbf{Irrelevant}: irrelevant revision is about examples or details that are appropriate and unnecessary, impertinent to, or disconnected from claims.
\item \textbf{Repeated}: repeat evidence revision is about examples or details that already exist.
\item \textbf{LCE}: linked claim-evidence revision is about the explanation that connects evidence with claims. 
\item \textbf{Not LCE}:  non-linked claim-evidence revision is about the explanation that does not connect evidence with claims.
\item \textbf{Commentary}: commentary revision is about the explanation that is unrelated to claims or source text; instead, it is the writer’s personal experience.
\item \textbf{Other}: other argument revisions that are not included in the above intentions.
\end{itemize}

We show annotated sentence-level revision examples in Table~\ref{table:data-example-argrevision}, where the original sentence is empty in adding, and the revised sentence is empty in deleting.

\subsection{ITERATER Examples}
\label{sec:iterater_example}
We used a publicly available annotated corpus named ITERATER (Du et al. 2022b), which contains revisions annotated from Wikipedia, ArXiv, and news articles. Wikipedia revisions typically aim to enhance clarity and structure. ArXiv edits are made by researchers who revise content, such as hypotheses, experimental findings, and interpretations. News article edits are performed by editors who focus on improving clarity and readability. The annotation taxonomy includes:

\begin{itemize}
[leftmargin=*,itemsep=-5pt, topsep=2pt]
\item \textbf{Fluency}: the revision to fix grammatical errors in the text.
\item \textbf{Coherence}: the revision to make the text more cohesive, logically linked, and consistent.
\item \textbf{Clarity}: the revision to make the text more formal, concise,  readable, and understandable.
\item \textbf{Style}: the revision to convey the writer’s writing preferences,  including emotions, tone, voice, etc.
\item \textbf{Meaning-Changed}: the revision to update or add new information to the text.
\item \textbf{Other}: the revisions that are not recognizable and do not belong to the above intentions.
\end{itemize}

The annotated ITERATER corpus includes ITERATER-sent and ITERATER-doc datasets. We include annotated sentence-level revision examples in Table~\ref{table:data-example-iterater}.

\begin{table}[!t]\small
\centering
\begin{adjustbox}{width=0.98\columnwidth}
\begin{tabular}{p{0.45\textwidth}}
\toprule
        \#\#\# Instruction: revise the original text based on the intention(s). \\
        
        \vspace{0.001pt}
        Relevant: relevant revision is about examples or details that are relevant to the claim. \\
        Irrelevant: irrelevant revision is about examples or details that are appropriate and unnecessary, impertinent to, or disconnected from claims.\\
        Repeated: repeat evidence revision is about examples or details that already exist.\\
        LCE: linked claim-evidence revision is about the explanation that connects evidence with claims. \\
        Not LCE:  non-linked claim-evidence revision is about the explanation that does not connect evidence with claims.\\
        Commentary: commentary revision is about the explanation that is unrelated to claims or source text; instead, it is the writer’s personal experience.\\

        \vspace{0.001pt}

         Here is the example of the revision: \{example\} \\
         \vspace{0.001pt}

        \#\#\# Original text: \{original text\}\\
        \#\#\# Revised text: \\
\bottomrule
\end{tabular}
\end{adjustbox}
\vspace{-.1in}
\caption{The in-context learning prompt regarding six intentions in ArgRevision. }
\label{table:prompt-icl-argrevision}
\vspace{-0.15in}
\end{table}

\begin{table}[H]\small
\centering
\begin{adjustbox}{width=0.98\columnwidth}
\begin{tabular}{p{0.45\textwidth}}
\toprule
        \#\#\# Instruction: revise the original text based on the intention(s). \\
        
        \vspace{0.001pt}
        Relevant: relevant revision is about examples or details that are relevant to the claim. \\
        Irrelevant: irrelevant revision is about examples or details that are appropriate and unnecessary, impertinent to, or disconnected from claims.\\
        Repeated: repeat evidence revision is about examples or details that already exist.\\
        LCE: linked claim-evidence revision is about the explanation that connects evidence with claims. \\
        Not LCE:  non-linked claim-evidence revision is about the explanation that does not connect evidence with claims.\\
        Commentary: commentary revision is about the explanation that is unrelated to claims or source text; instead, it is the writer’s personal experience.\\
        \vspace{0.001pt}

        When you revise the text, please break down the revision process into multiple steps, and check if the revision fulfills the intention requirement in each step. Only output the final revised text. Do not output the intermediate steps. \\ \vspace{0.001pt}

         Here is the example of the revision: \{example\} \\
         \vspace{0.001pt}

        \#\#\# Original text: \{original text\}\\
        \#\#\# Revised text: \\
\bottomrule
\end{tabular}
\end{adjustbox}
\vspace{-.1in}
\caption{The in-context learning with chain-of-thought prompt regarding six intentions in ArgRevision.}
\label{table:prompt-cot-argrevision}
\vspace{-0.1in}
\end{table}

\subsection{In-Context Learning Prompts}
\label{sec:in-context-learning}
We design an in-context learning prompt based on~\citet{raheja-etal-2023-coedit} and~\citet{ziegenbein-etal-2024-llm} as a non-finetuned LLM baseline. The prompt includes the intention taxonomies described in Appendices~\ref{sec:argrevision_example} and~\ref{sec:iterater_example} to indicate the revision objectives\footnote{The intention taxonomies included in a prompt are determined by the intention labels associated with each revision. Not all the taxonomies are used if a prompt does not require the revisions to achieve all the intentions.}. The prompt also uses a one-shot example retrieved from the training set to demonstrate the revisions. Specifically, the original texts in the test and training sets are encoded with Sentence Transformer~\cite{thakur-2020-AugSBERT}. The top similar example is retrieved using Faiss~\cite{johnson2019billion}. The retrieved original text paired with its revised text is used as a one-shot example. In addition, we use a chain-of-thought prompt as another baseline, which adds the text ``\textit{when you revise the text, please break down the revision process into multiple steps, and check if the revision fulfills the intention requirement in each step. Only output the final revised text. Do not output the intermediate steps''} to the prompt. 
The detailed prompts are shown in Tables~\ref{table:prompt-icl-argrevision} and~\ref{table:prompt-cot-argrevision}, and Tables~\ref{table:prompt-icl-iterater} and~\ref{table:prompt-cot-iterater} for ArgRevision and ITERATER  datasets, respectively.

\begin{table}[!tb]\small
\centering
\begin{adjustbox}{width=0.98\columnwidth}
\begin{tabular}{p{0.45\textwidth}}
\toprule
        \#\#\# Instruction: revise the original text based on the intention(s). \\
        
        \vspace{0.001pt}
        Fluency: the revision to fix grammatical errors in the text. \\
        Coherence: the revision to make the text more cohesive, logically linked, and consistent. \\
        Clarity: the revision to make the text more formal, concise, readable, and understandable. \\
        Style: the revision to convey the writer’s writing preferences, including emotions, tone, voice, etc. \\
        Meaning-Changed: the revision to update or add new information to the text. \\
        \vspace{0.001pt}

         Here is the example of the revision: \{example\} \\
         \vspace{0.001pt}

        \#\#\# Original text: \{original text\}\\
        \#\#\# Revised text: \\
\bottomrule
\end{tabular}
\end{adjustbox}
\vspace{-.1in}
\caption{The in-context learning prompt regarding five intentions in ITERATER. }
\label{table:prompt-icl-iterater}
\vspace{-0.05in}
\end{table}

\begin{table}[!tb]\small
\centering
\begin{adjustbox}{width=0.98\columnwidth}
\begin{tabular}{p{0.45\textwidth}}
\toprule
        \#\#\# Instruction: revise the original text based on the intention(s). \\
        
        \vspace{0.001pt}
        Fluency: the revision to fix grammatical errors in the text. \\
        Coherence: the revision to make the text more cohesive, logically linked, and consistent. \\
        Clarity: the revision to make the text more formal, concise, readable, and understandable. \\
        Style: the revision to convey the writer’s writing preferences, including emotions, tone, voice, etc. \\
        Meaning-Changed: the revision to update or add new information to the text. \\
        \vspace{0.001pt}

        When you revise the text, please break down the revision process into multiple steps, and check if the revision fulfills the intention requirement in each step. Only output the final revised text. Do not output the intermediate steps. \\ \vspace{0.001pt}

         Here is the example of the revision: \{example\} \\
         \vspace{0.001pt}

        \#\#\# Original text: \{original text\}\\
        \#\#\# Revised text: \\
\bottomrule
\end{tabular}
\end{adjustbox}
\vspace{-.1in}
\caption{The in-context learning with chain-of-thought prompt regarding five intentions in ITERATER.}
\label{table:prompt-cot-iterater}
\vspace{-0.19in}
\end{table}

\subsection{Hyperparameters and Metrics}
\label{sec:hyperparamter}
We implement the framework using Python v3.9.20, PyTorch v2.5.1, and HuggingFace Transformers v4.46.0, and utilize Weights \& Biases v0.19.1 for logging. The used fine-tuning hyperparameters\footnote{The values of the \textit{max length} parameter are doubled in the ICL and CoT baselines (i.e., setting \textit{max length} as 512, 2,048, 3,072) since their prompts include one-shot examples that are long documents exceeding the limits in PEFT settings.} are shown in Table~\ref{tab:parameter}. We use SARI\footnote{\href{https://github.com/cocoxu/simplification}{https://github.com/cocoxu/simplification}}, GLEU\footnote{\href{https://github.com/facebookresearch/EditEval}{https://github.com/facebookresearch/EditEval}}, and Update-R\footnotemark[6] as revision generation metrics. Here, SARI and Update-R are crucial metrics for text revision, given that the original and revised texts often exhibit high overlap in words. We use evaluation packages adopted from prior work~\cite{dwivedi-yu-etal-2024-editeval,xu2016optimizing}.

\begin{table}[tb]
\begin{adjustbox}{width=1\columnwidth}
\begin{tabular}{c|ccc}
\toprule
Hyperparameter & ITERATER-sent & ITERATER-doc & ArgRevision \\ \midrule
rank $r$ & \multicolumn{3}{c}{32} \\
$\alpha$ & \multicolumn{3}{c}{64} \\
dropout & \multicolumn{3}{c}{0.05} \\
where & \multicolumn{3}{c}{\{Q, K, V, Up, Down\}} \\
fine-tuning batch size & 16 & 4 \textbf{{\space\space\space\space\space}} & 2 \\
inference batch size & 32 &8 \textbf{{\space\space\space\space\space}} & 4 \\
max length & 256 &1024 \textbf{{\space\space\space\space\space}} & 1536 \\
max new tokens & 128 &512 \textbf{{\space\space\space\space\space}} & 768 \\
optimizer & \multicolumn{3}{c}{AdamW} \\
learning rate & \multicolumn{3}{c}{2e-4} \\
lr scheduler & \multicolumn{3}{c}{Warmup Steps} \\
weight decay & \multicolumn{3}{c}{0} \\
warmup steps & \multicolumn{3}{c}{100} \\
epochs & \multicolumn{3}{c}{2} \\
sample & \multicolumn{3}{c}{true} \\
beam num & \multicolumn{3}{c}{4} \\
top p & \multicolumn{3}{c}{0.75} \\
top k & \multicolumn{3}{c}{40} \\
temperature & \multicolumn{3}{c}{0.2} \\
\bottomrule
\end{tabular}
\end{adjustbox}
\caption{Hyperparameters for PEFT with Intention-Tuning on ITERATER-sent, ITERATER-doc, and ArgRevision.}
\label{tab:parameter}
\vspace{-.2in}
\end{table}

\begin{figure*}[tb]
    \centering
    \includegraphics[width=2.09\columnwidth]{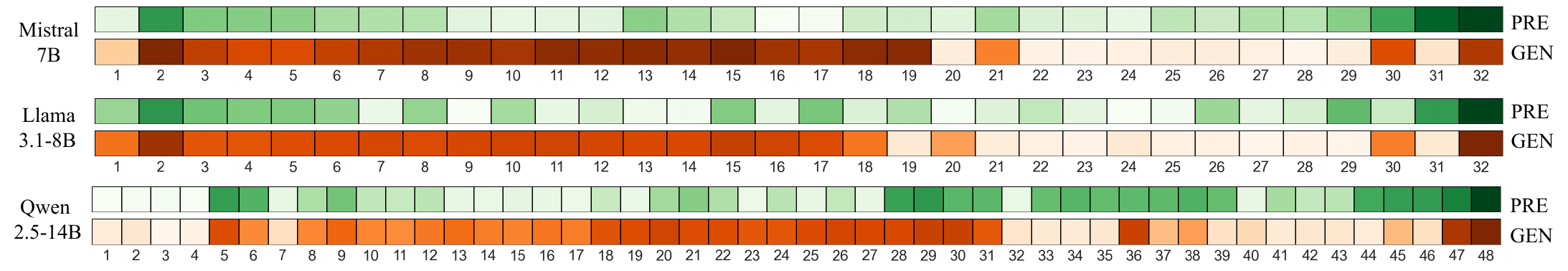}
    \vspace{-2.0em}
    \caption{The layer-wise importance score alignment between the intention prediction (green boxes) and the revision generation (brown boxes) tasks while fine-tuning LLMs on the ITERATER-doc. All PEFT uses LoRA. The dark colors indicate high scores (important layers) while the light colors indicate low scores (redundant layers).}
    \vspace{-.2em}
    \label{fig:layer-gradient-iterater-doc}
\end{figure*}

\begin{figure*}[tb]
    \centering
    \includegraphics[width=2.09\columnwidth]{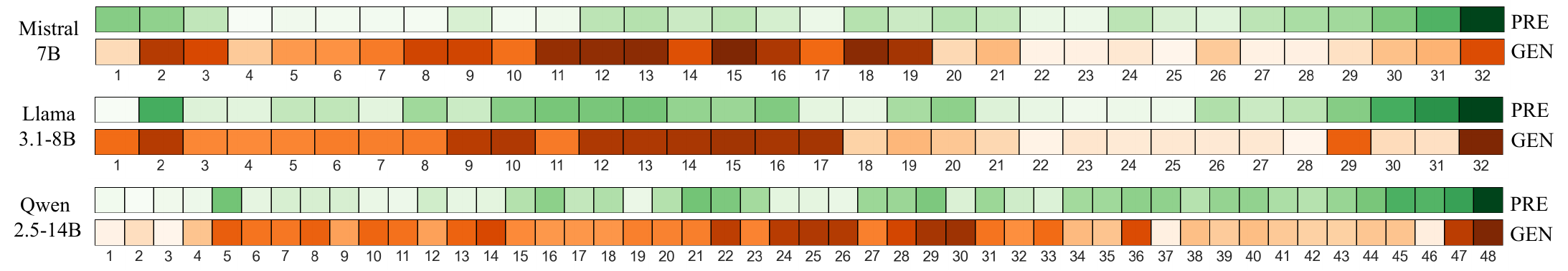}
    \vspace{-2.0em}
    \caption{The layer-wise importance score alignment between the intention prediction (green boxes) and the revision generation (brown boxes) tasks while fine-tuning LLMs on the ArgRevision. All PEFT uses LoRA. The dark colors indicate high scores (important layers) while the light colors indicate low scores (redundant layers).}
    \vspace{-.5em}
    \label{fig:layer-gradient-argrevision}
\end{figure*}

\subsection{Layer Alignment}

We visualize the LLM layer alignment across three LLMs in Figure~\ref{fig:layer-gradient-iterater-doc} and Figure~\ref{fig:layer-gradient-argrevision}, specifically for the multi-intent revision datasets. Despite architectural differences, the important and redundant layers show observable alignment between the intention prediction and the revision generation tasks. Regarding ITERATER-doc, the dark green and dark brown boxes coincide, especially for layers 2 and 32 in Mistral-7B and Llama3.1-8B, and layers 5 to 12 and 28 to 31 in Qwen2.5-14B. In terms of ArgRevision, the alignments are relatively higher on Llama3.1-8B than Mistral-7B and Qwen2.5-14B (see Table~\ref{tab:importance-score-correlation}). Although layer alignment varies across LLMs, the important layers used for revision generation remain consistent across datasets, e.g., layers 5 to 31 in Qwen2.5-14B. This observation highlights that specific LLM layers are consistently important in revision generation tasks across different datasets; however, this does not hold true for intention prediction, which might be because LLMs are largely pre-trained for generation rather than classification. This observation will be evaluated in future work.

\subsection{Convergence on Adapters}
\label{sec:convergence_adapter}
\begin{figure*}[tb]
    \centering
    \includegraphics[width=1.7\columnwidth]    {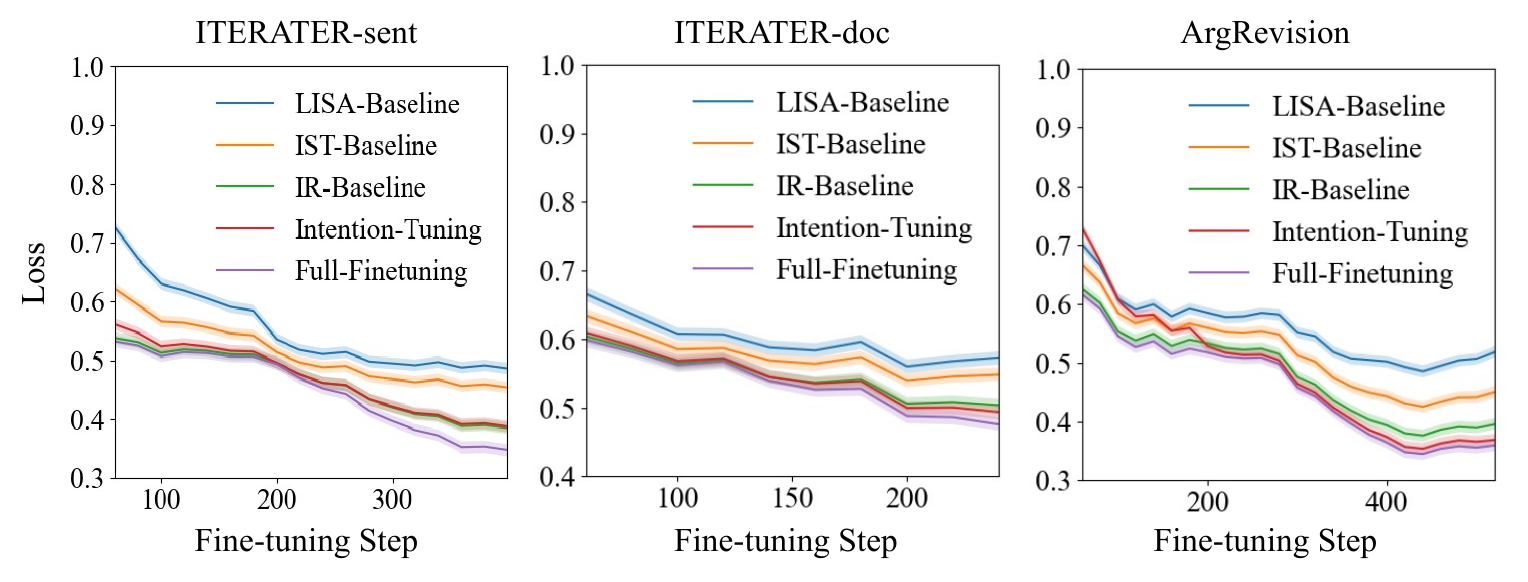}
    \caption{Llama3.1-8B fine-tuning loss on the training set of the three datasets. All PEFT uses DoRA.}
    \label{fig:fine-tuning-loss-dora}
\end{figure*}

\begin{figure*}[tb]
    \centering
    \vspace{-.1in}
    \includegraphics[width=1.7\columnwidth]    {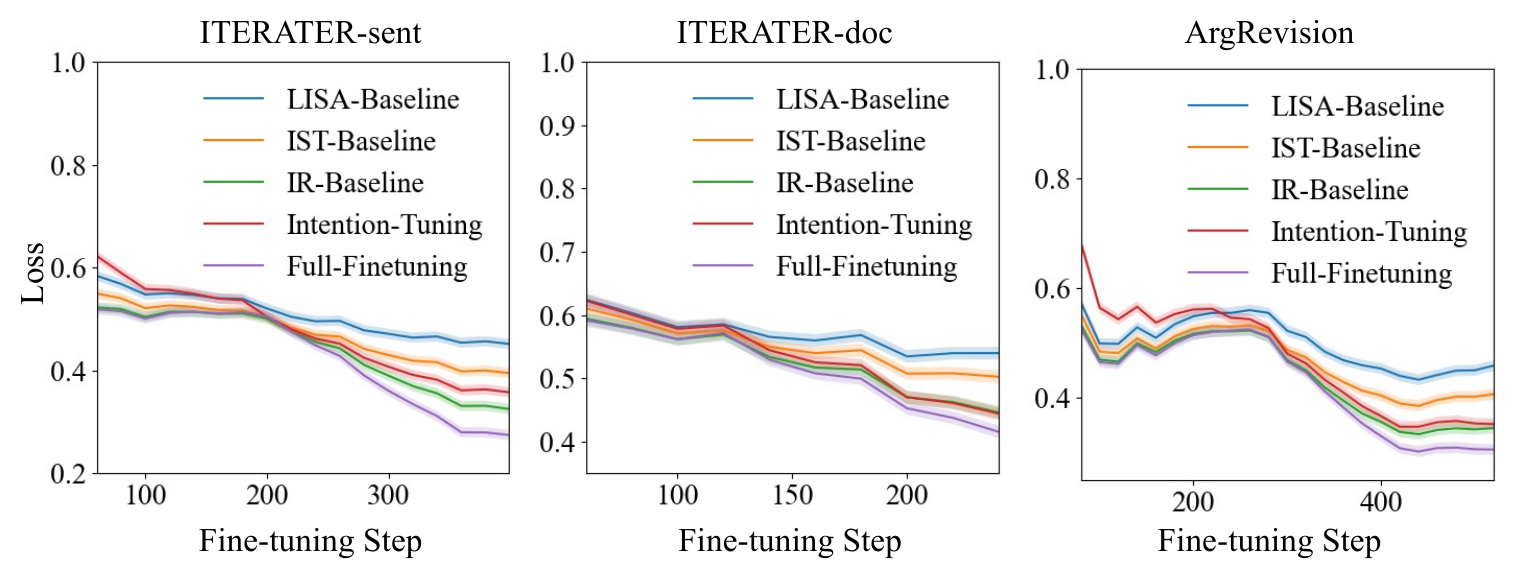}
    \caption{Llama3.1-8B fine-tuning loss on the training set of the three datasets. All PEFT uses Bottleneck.}
    \label{fig:fine-tuning-loss-bottleneck}
\end{figure*}

We conduct additional experiments to visualize the loss convergence of Llama3.1-8B while fine-tuning with DoRA~\cite{pmlr-v235-liu24bn} and Bottleneck~\cite{houlsby2019parameter} adapters. Figure~\ref{fig:fine-tuning-loss-dora} and~\ref{fig:fine-tuning-loss-bottleneck} show that Intention-Tuning has fast convergence across different adapters and datasets. Although the loss convergence in IR-Baseline is more rapid than Intention-Tuning, they are close. These observations suggest that Intention-Tuning is generally efficient during fine-tuning.

\subsection{Error Analysis}
\label{sec:error_analysis}
\begin{figure}[tb]
    \centering
    \includegraphics[width=1\columnwidth]{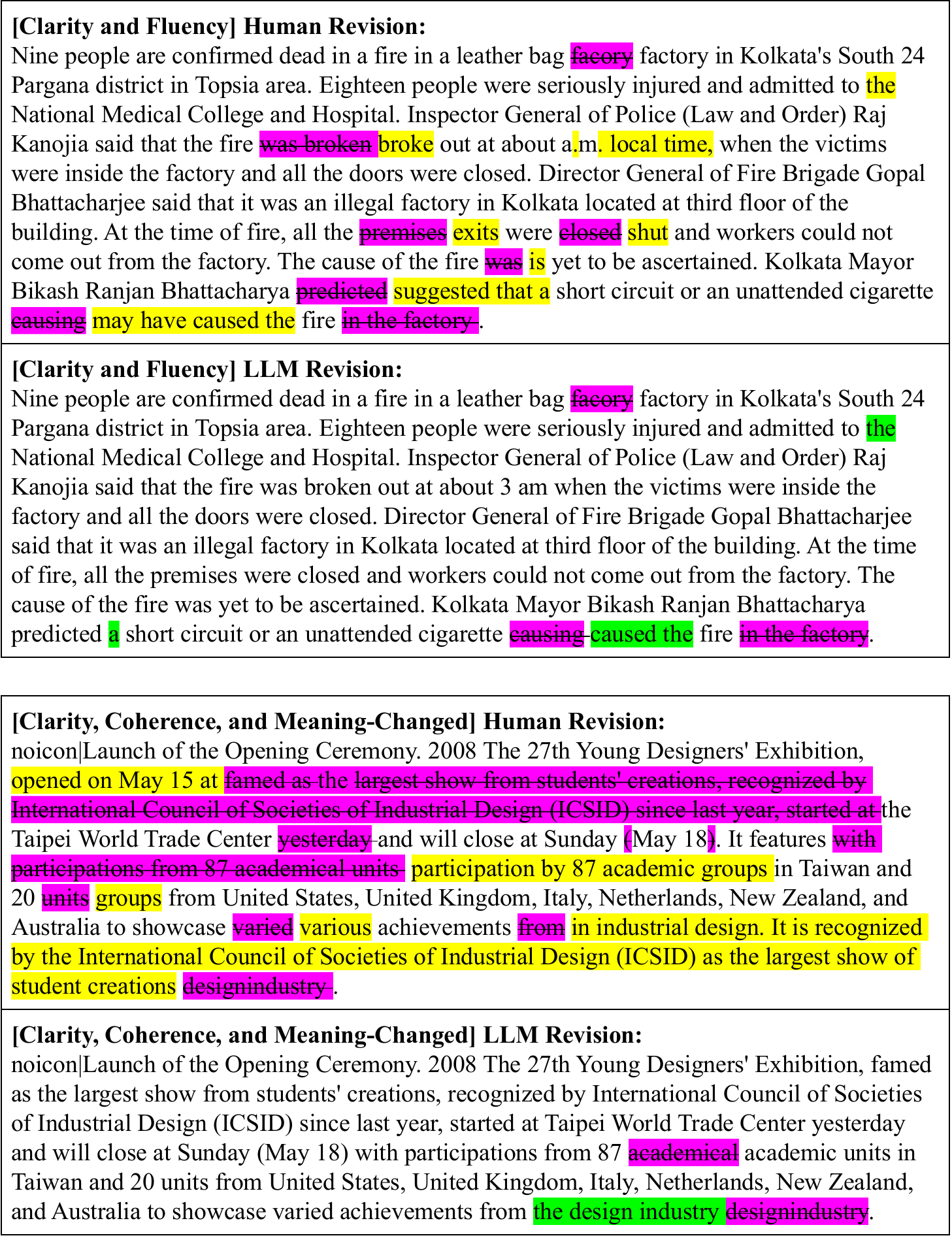}
    \caption{The example of the human and generated revisions by Llama3.1-8B with Intention-Tuning on the ITERATER-doc dataset. The yellow and green colors denote additions, and the purple denotes deletions.}
    \label{fig:example-human-llm-iterater-doc}
\end{figure}

We analyze multi-intent revision generation on the ITERATER-doc and ArgRevision datasets. Figure~\ref{fig:example-human-llm-iterater-doc} shows \textit{clarity} and \textit{fluency} revisions mostly involve wording and surface issues. The human replaces ``was broken'' with ``broke'', and ``closed'' with ``shut''; however, the LLM does not make these revisions. Instead, it corrects the typo ``facory'' with ``factory'', and replaces ``causing'' with ``caused'' to improve \textit{clarity}, which is in line with human edits. This example illustrates that LLMs can generate desirable revisions to align with the writer's intentions, but have limited capability to identify the text spans that have such issues, resulting in minimal revisions. Similarly, another example shows that human involves more content-level revisions, e.g., adding additional information to indicate the exhibition ``opened on May 15'', and revises sentence orders to improve \textit{coherence}. In contrast, the LLM is limited in handling such revisions. Although LLMs with Intention-Tuning achieve relatively higher performance than the baselines, they are more conservative than humans in making revisions, which could contribute to counter-intuitive high GLEU and Average scores, similar to the Copy-Baseline performance in Table~\ref{tab:peft-comparision}.

\begin{figure}[tb]
    \centering
    \includegraphics[width=1\columnwidth]{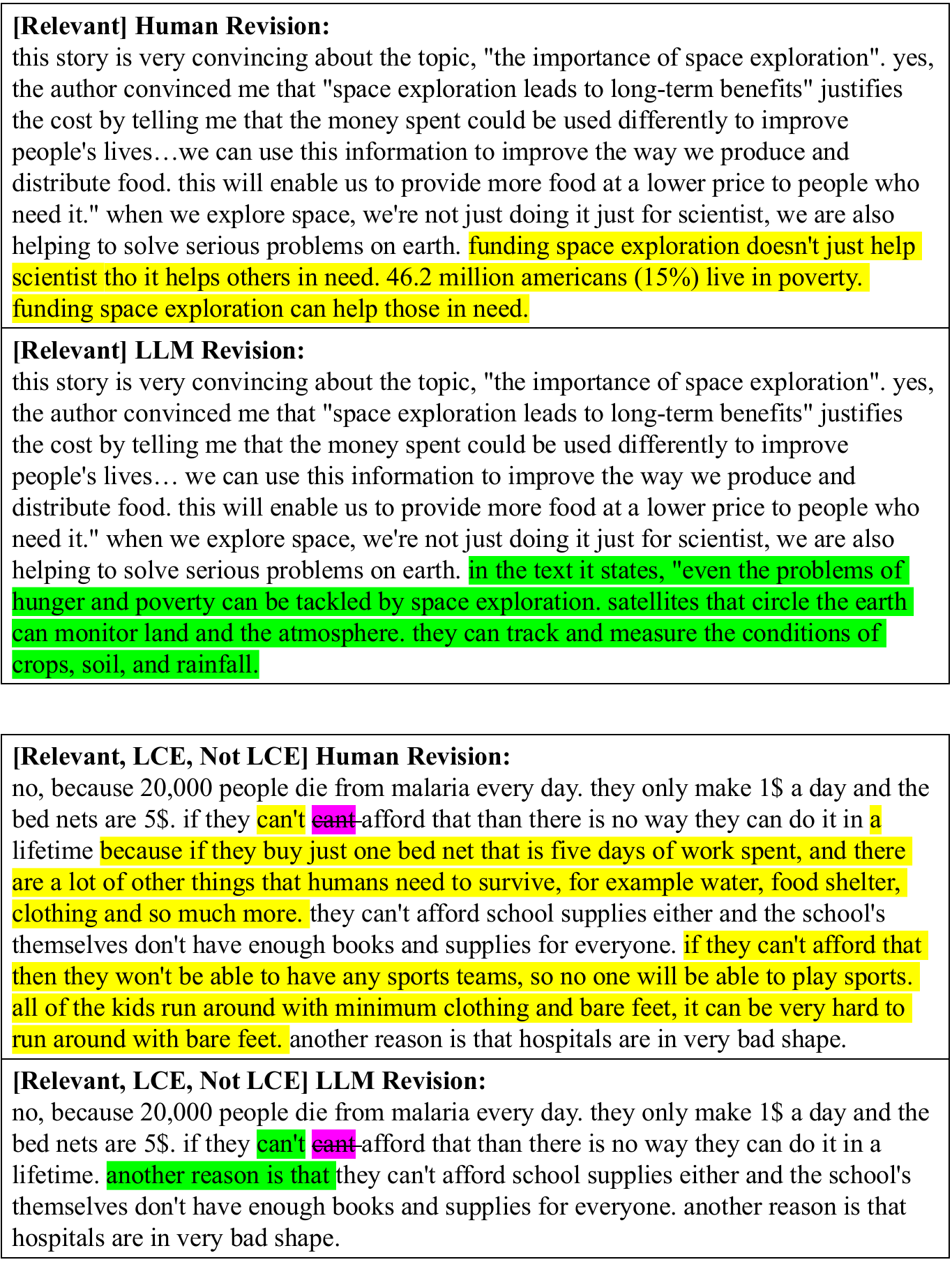}
    \caption{The example of the human and generated revisions by Llmam3.1-8B with Intention-Tuning on the ArgRevision dataset. The yellow and green colors denote additions, and the purple denotes deletions.}
    \label{fig:example-human-llm-erevise}
\end{figure}

Figure~\ref{fig:example-human-llm-erevise}
shows argument revisions in ArgRevision. The human adds relevant examples to explain that the reason why they need ``funding space exploration'' is not only because it helps the ``scientist,'' but also because it “helps others.” Although the LLM revision provides evidence to support the claim, its content does not align with the human revision. These observations illustrate the challenges in argument revision generation, which requires providing relevant evidence linked to the main argument. In addition, another example shows the LLM makes minimal changes, despite being requested for \textit{relevant}, \textit{LCE}, and \textit{not LCE} revisions. This again demonstrates that LLMs are conservative in providing challenging revisions, possibly because they struggle to identify the text spans (e.g., sentences that require revision) and are uncertain about revision actions (e.g., adding, deleting, or modifying sentences). These examples offer insights into enhancing revision quality by improving the identification of revision spans and actions in future work.

\begin{table*}[th]
\centering
\begin{adjustbox}{width=1.95\columnwidth}
\begin{tabular}
{>{\centering\arraybackslash}m{0.03\textwidth}m{0.35\textwidth}m{0.35\textwidth}>{\centering\arraybackslash}m{0.1\textwidth}m{0.08\textwidth}}	
\toprule
ID & Sentence in original text & Sentence in revised text & Edit & Intention \\ \midrule
1 & I think that we should still help nasa but keep in mind that there are lots of kids and adults out there that need help. & I think that we should still help nasa but keep in mind that there are lots of kids and adults out there that need help. & N/A & N/A \\ \midrule
2 & ... & ... & ... \\ \midrule
3 &  & Nineteen billion was only 1.2\% of the total national budget. & Add & Irrelevant \\ \midrule
4 & So i think that we should keep giving money but keep thinking of others. &  & Delete & not LCE \\ \midrule
5 &  & It is very important to keep thinking of others. & Add & LCE \\ \midrule
6 & Because it is not just people that need help. & because it is not just people that need help. & N/A & N/A \\ \midrule
7 & ...  & ... & ... \\ \midrule
8 & A program to develop clean energy could be viewed as a worthy investment. & A program to develop clean energy could be viewed as a worthy investment. & N/A & N/A \\ \midrule
9 & Maybe exploring space should not be a priority when there is so much that needs to be done on earth. & Maybe exploring space should not be a priority when there is so much that needs to be done on earth. & N/A &  N/A \\ \midrule
10 & In 2012, united states spent 19 billion dollars for space exploration. & In 2012, united states spent 19 billion dollars for space exploration. & N/A & N/A \\ \midrule
11 & Some people think that money should be spent to help heal the people and the earth. & Some people think that money should be spent to help heal the people and the earth. & N/A & N/A \\ \midrule
12 &  & The arguments against space exploration stem from a belief that the money spent could be used differently – to improve people’s lives. & Add & Relevant \\ \midrule
13 &  & In 1953, president eisenhower captured this viewpoint.& Add & Relevant \\ \midrule
14 & ... & ... & ... \\ \midrule
15 & Also nineteen billion was only 1.2\% of the total national budget. &  & Delete & Irrelevant \\\bottomrule
\end{tabular}
\end{adjustbox}
\caption{Example of revision intention annotation for an essay in ArgRevision.}
\label{table:data-example-argrevision}
\end{table*}

\begin{table*}[!htb]\small
\centering
\begin{adjustbox}{width=1.75\columnwidth}
\begin{tabular}
{>{\centering\arraybackslash}m{0.01\textwidth}m{0.27\textwidth}m{0.27\textwidth}>{\centering\arraybackslash}m{0.12\textwidth}}	
\toprule

ID & Sentence in original text & Sentence in revised text & Intention \\ \midrule
1 & Bidirectional Encoder Representations from Transformers (BERT) models for biomedical specialties such as BioBERT and clinicalBERT have significantly improved in biomedical text-mining tasks and enabled us to extract valuable information from biomedical literature . & Bidirectional Encoder Representations from Transformers (BERT) models for medical specialties, such as BioBERT and clinicalBERT have significantly improved in biomedical text-mining tasks and enabled us to extract valuable information from biomedical literature . & Style \\ \midrule
2 & We introduce the method to train a BERT model on a small medical corpus both in English and Japanese, respectively, and then we evaluate each of them in terms of the biomedical language understanding evaluation (BLUE) benchmark and the medical-document-classification task in Japanese, respectively. & We introduce the method to train a BERT model on a small medical corpus both in English and Japanese, respectively, and then we evaluate each of them in terms of the biomedical language understanding evaluation (BLUE) benchmark and the medical document classification task in Japanese, respectively. & Clarity \\ \midrule
3 & Therefore, we propose a method that realizes a high-performance BERT model by using a small corpus. & Therefore, we propose a method that realizes a high-performance BERT model using a small corpus. & Coherence \\ \midrule
4 & After confirming their satisfactory performances, we apply our method to develop a model that outperforms the pre-existing models. & After confirming their satisfactory performances, we applied our method to develop a model that outperforms the pre-existing models. & Fluency \\ \midrule
5 & The total score is 1.0 points above that of BioBERT . & The total score is 1.0 points above that of BioBERT and 0.3 points above that of the ablated model trained without our proposed method. This proposed technique is an effective approach to develop localized medical BERT models and to enhance domain-specific models in the biomedical domain . & Meaning-Changed \\

\bottomrule
\end{tabular}
\end{adjustbox}
\caption{Example of revision intention annotation for an article in ITERATER.}
\label{table:data-example-iterater}
\vspace{2in}
\end{table*}

\end{document}